\documentclass[10pt,journal,compsoc]{IEEEtran}

\usepackage{xcolor,soul,framed} %,caption

\colorlet{shadecolor}{yellow}
\usepackage[pdftex]{graphicx}
\graphicspath{{../pdf/}{../jpeg/}}
\DeclareGraphicsExtensions{.pdf,.jpeg,.png}

\usepackage[cmex10]{amsmath}
\usepackage{array}
\usepackage{mdwmath}
\usepackage{mdwtab}
\usepackage{eqparbox}
\usepackage{url}
\usepackage{graphicx}
\usepackage{amsmath}
\usepackage{amsfonts}
\usepackage{algpseudocode}
\usepackage{multirow}
\usepackage{footnote}
\usepackage{bm}
\usepackage{makecell}
\usepackage{pdflscape}
\usepackage{afterpage}
\usepackage{bbding}
\usepackage{lscape}
\usepackage{booktabs}
\usepackage{tikz}
\usepackage{bbding}
\usepackage{pgfplots}
\usepackage{caption}
\usepackage{rotating}
\usepackage{tabularray}
\usepackage{tabularx}
\usepackage{makecell}
\usepackage{ragged2e}
\usepackage{threeparttable}
\usepackage{algorithm}
\usepackage{amssymb}

\newcolumntype{L}{>{\RaggedRight\hangafter=1\hangindent=0em}X}
\newcolumntype{C}{>{\Centering\hangafter=1\hangindent=0em}X}

\pgfplotsset{compat=1.12}
\usepackage{subcaption}
    \pgfplotsset{
        layers/my layer set/.define layer set={
            background,
            main,
            foreground
        }{ },
        set layers=my layer set,
    }

\usepackage[square,sort,comma,numbers]{natbib} % multiple citation or 
\usepackage[top=0.6in,bottom=0.4in,left=0.4in,textwidth=7.7in]{geometry}
\usepackage[pagebackref=false,breaklinks=false,linkcolor=red,anchorcolor=black, citecolor=green,colorlinks,bookmarks=true]{hyperref}
\usepackage{amsfonts}

\usepackage{pifont}

\usepackage{pifont} % 在导言区引入宏包
\newcommand{\xmark}{\ding{55}} % 定义快捷命令

\newcommand{\tabincell}[2]{\begin{tabular}{@{}#1@{}}#2\end{tabular}} 
\begin{document}

\title{Toward Active Object Detection for UAVs in the Wild: \\A Large-Scale Dataset, Benchmark and Method}

\author{Tianpeng Liu, Xinhua Jiang, Li Liu, Qinmu Shen, Siwei Tang, Zhen Liu, Yongxiang Liu  
        % <-this % stops a space
\thanks{
This work was supported by National Natural Science Foundation of China (NSFC) under Grant Nos. 62376283 and 62531026; by the Fundamental and Interdisciplinary Disciplines Breakthrough Plan of the Ministry of Education of China under Grant JYB2025XDXM110; and by the Innovation Research Foundation of National University of Defense Technology under Grant JS2023-03.
\emph{($^{\ast}$Tianpeng Liu and Xinhua Jiang are are co-first authors.)}
\emph{($^{\ast}$Corresponding authors: Li Liu and Yongxiang Liu. e-mail: liuli\_nudt@nudt.edu.cn and  lyx\_bible@sina.com.)}}% <-this % stops a space
\thanks{All the authors are with the College of Electronic Science and Technology, National University of Defense Technology, Changsha, 410073, China.}
}

\markboth{Submitting to IEEE TPAMI}%
{Liu \MakeLowercase{\textit{et al.}}: ATRNet-LUDO}

\IEEEtitleabstractindextext{%

\begin{abstract}
\justifying 
Object detection is a fundamental component in numerous Unmanned Aerial Vehicle (UAV) applications, yet it has long been plagued by hindrances like occlusion or target pixel scarcity. Active Object Detection (AOD) provides a novel paradigm to address these challenges via active vision, while UAV-based AOD research remains scarce due to the lack of high-quality datasets and benchmarks for algorithm development and evaluation. To fill this gap, this paper presents \textbf{ATRNet-LUDO}, the first large-scale real-world dataset for UAV-Ground Active Object Detection (UGAOD). It contains 121,000 multi-view panoramic multi-target aerial images and 1.21 million local single-target slices, covering 10 vehicle targets across 40 scenarios. It enables the construction of diverse training and testing environments for UAV agent interaction and active observation policy learning. Based on this dataset, we establish a comprehensive evaluation benchmark for AOD policy learning methods. Most existing AOD policies rely on Deep Reinforcement Learning (DRL) but suffer from poor generalization. Evaluations on our benchmark reveal a significant generalization gap between training and testing performance, highlighting an urgent need for solutions. To this end, we leverage the Joint Embedding Predictive Architecture (JEPA) to construct a world model that enhances state representation learning, and propose AOD-JEPA by incorporating AOD-specific prior knowledge. Extensive experiments validate its effectiveness: active observation policies outperform traditional passive perception by ~20 percentage points in target recognition rate, and the proposed world model-aided active observation policy achieves a 2–3 percentage recognition gain over DRL baselines in test environments with similar UAV movement costs. We hope ATRNet-LUDO and the benchmark will advance research in the UGAOD field. The dataset and code are soon available at \url{https://github.com/Leo000ooo/LUDO_dataset}.

\end{abstract}

} 

\maketitle
\IEEEdisplaynontitleabstractindextext
\IEEEpeerreviewmaketitle

\IEEEraisesectionheading{\section{Introduction}
\label{Introduction}}

\IEEEPARstart{U}{nmanned} Aerial Vehicles (UAVs) have emerged as indispensable tools across diverse domains, including smart agriculture \cite{1}, aerial surveying and mapping \cite{2}, disaster rescue \cite{3,4}, and traffic monitoring \cite{5,6}, owing to their distinct advantages of flexible operation, wide field of view, and high maneuverability \cite{33}. Underpinning these practical applications is the target detection technology based on UAV-borne imagery \cite{7}, the core objective of which is to accurately locate and identify objects of interest within the captured images. Nevertheless, complex ground objects surrounding the target often inevitably introduce interference to the target detection process \cite{8,9}. Existing occlusion-resistant target detection methods merely involve adaptive modifications to the model architecture, and their performance still falls short of the desired standards \cite{10}. In contrast, Active Object Detection (AOD) integrates the paradigm of active vision \cite{11}, enabling fundamental enhancements to the quality of observed imagery and thereby yielding improved target detection performance \cite{13}. Specifically, the AOD task is defined as follows: Given an initial image and the approximate location of the target to be identified, the decision-making module of the embodied agent performs path planning in the environment using historical observation data. The goal is to move to the optimal observation position with minimal time cost, enabling the perception module to accurately recognize the target from the final observed image, so as to mitigate the adverse effects caused by environmental factors such as target occlusion.

\IEEEpubidadjcol

\begin{figure*}[!t]
	\centering
	\includegraphics[width=0.8\linewidth]{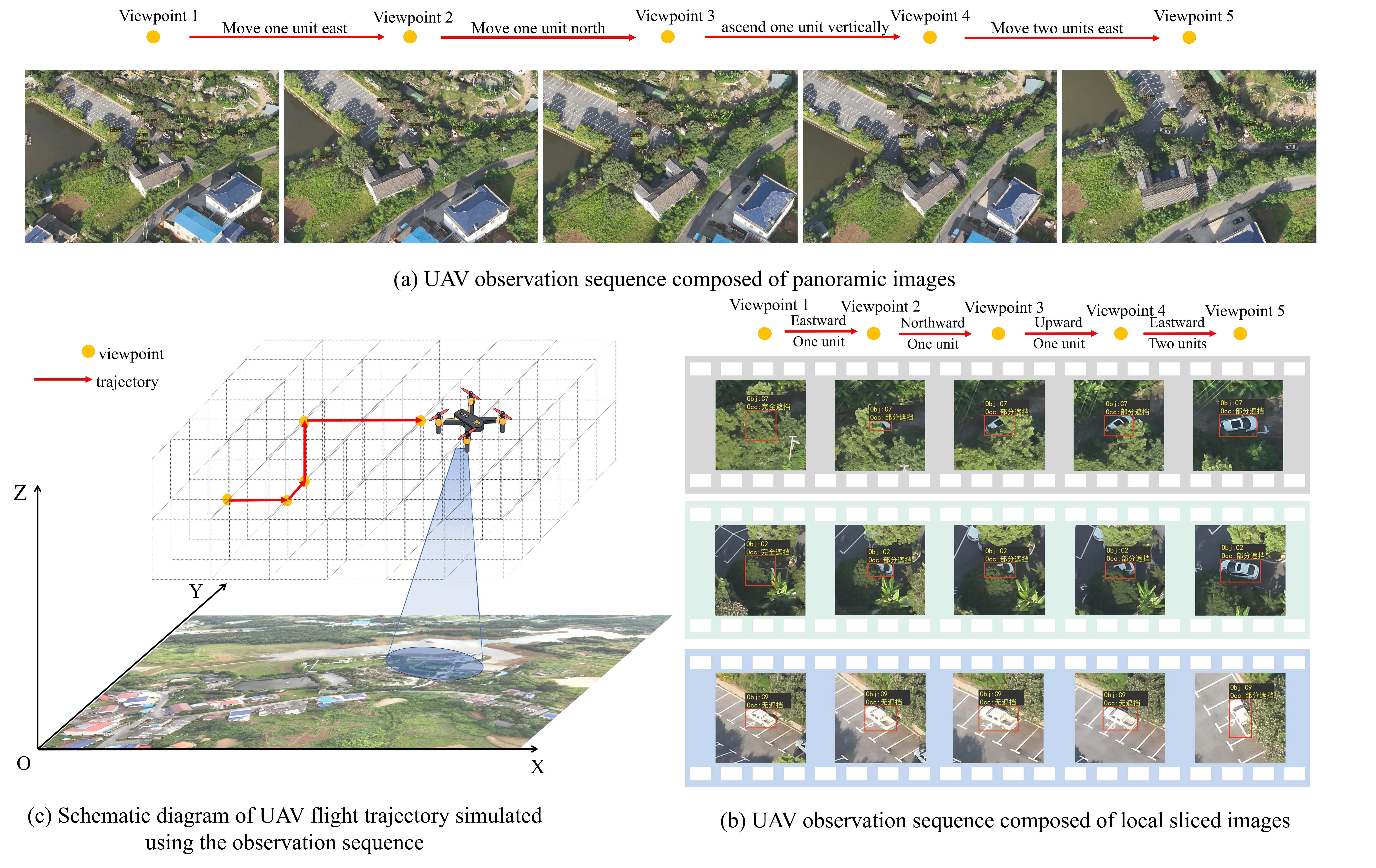}
	\caption{Illustrations of sample instances and usage protocols for the ATRNet-LUDO dataset. (a) The observation sequence of multi-view panoramic images collected by the UAV during its flight along the trajectory shown in subfigure (c). (b) The observation sequence of multi-view local target patches collected by the UAV during its flight along the trajectory shown in subfigure (c). (c) Schematic diagram of a UAV flying along a specific trajectory; we define UAV flight trajectories based on combinations of preset discrete sampling points. Observational images are collected by the UAV at these sampling points, and thus sequences of panoramic images or local patches from adjacent viewpoints can simulate the results of continuous observations made by the UAV during flight.}
	\label{fig_1}
\end{figure*}

%The AOD task is a perceptual task proposed for embodied agents. An embodied agent typically comprises three core modules: a perception module responsible for object detection and recognition, a decision-making module for agent path planning, and a motion control module that governs the agent’s movement based on output actions. 

The advancement of embodied intelligence is intrinsically reliant on the construction of training and test environments. As an embodied perceptual task, the UAV-ground Active Object Detection (UGAOD) task also demands environments capable of providing interactive experiences for the agent. Unfortunately, current research in the AOD domain has predominantly focused on indoor scenarios \cite{13,21,28,29,30,31,32,39}, while research on the UGAOD remains largely unaddressed. The primary rationale for this gap is the lack of datasets that can be leveraged to construct training and test environments. Hence, this paper aims to construct real-scene-sampled environments tailored to UGAOD. The corresponding dataset must satisfy characteristics including multi-scenario coverage, diverse target categories, and dense air-to-ground multi-view sampling. Most existing UAV image datasets \cite{33,34,35,36} fail to meet these criteria. Only two public datasets—UEVAVD \cite{39} and CARLA-AOD \cite{21}—are relatively aligned with the requirements; however, both are sampled from virtual simulation environments, and suffer from insufficient overall data volume, leading to weak persuasiveness when used for benchmark evaluation.

%Environments can be classified into real-world environments and simulated environments based on the authenticity of interactions. While the former offers distinct advantages in terms of data authenticity, the interaction process between an agent and the real world is time-consuming and labor-intensive, entails potential safety risks, and hinders the establishment of benchmarks that can be rapidly reused by the academic community. In contrast, simulated environments feature low application costs, high safety, and strong scalability, facilitating the development of evaluation benchmarks under controlled experimental conditions \cite{14}. Simulated environments can be further subdivided into virtual simulation environments and real-scene-sampled simulation environments. The former are entirely constructed by mimicking the physical world within virtual spaces; examples include environments built with simulators such as Gazebo \cite{15}, Pybullet \cite{16}, and Mujoco \cite{17}. The latter, by contrast, primarily rely on data collected from the real world to build 3D simulated scenes, with representative simulators including AI2-ThOR \cite{18}, Matterport3D \cite{19}, and InfiniteWorld \cite{20}. 

This paper releases ATRNet-LUDO, the first large-scale real-world public dataset tailored for the UGAOD task. Here, "ATRNet" denotes the general name of the dataset series collected by our research group, while "LUDO" is an acronym for "a \textbf{L}arge-scale dataset for \textbf{U}AV active \textbf{O}bject \textbf{D}etection in the wild". The dataset comprises 121,000 multi-view panoramic multi-target aerial images and 1,210,000 multi-view local single-target aerial image slices, capturing 10 categories of vehicle targets across 40 distinct scenarios. Based on this dataset, we can construct diverse environments: 200 for multi-target AOD tasks and 2000 for single-target AOD tasks. In each environment, the combination of panoramic images or local target slices from adjacent viewpoints is regarded as the observation results obtained by the UAV flying along a specific trajectory. As illustrated in Fig.~\ref{fig_1}(a), the UAV observation sequence is composed of panoramic images from different viewpoints, with each frame containing multiple vehicle targets. Fig.~\ref{fig_1}(b) presents a sequence consisting of local patches focusing on individual vehicle targets in the panoramic images; these two types of sequences can be used to simulate the UAV's observational results of ground targets when flying along the trajectory shown in Fig.~\ref{fig_1}(c). Owing to the dense multi-view sampling principle adopted in the ATRNet-LUDO dataset, the combination of observational images can form abundant observation sequences, thereby enabling the simulation of a vast number of UAV flight trajectories to meet the research and validation requirements of UAV-ground single-target and multi-target AOD policy learning methods. Meanwhile, since the proposed dataset includes various occlusion obstacles and records the UAV's longitude and latitude, lens azimuth and pitch information, as well as the targets’ longitude and latitude during sampling, it can also be utilized to evaluate UAV image target detection, recognition, and localization algorithms under occluded conditions. Additionally, we establish an evaluation benchmark based on the ATRNet-LUDO dataset, which defines the evaluation tasks and metrics. On this basis, we evaluate 7 baseline AOD policy learning methods, aiming to provide an impartial and persuasive benchmark for the subsequent research and development in this field.

On the other hand, existing AOD policy learning methods primarily rely on Deep Reinforcement Learning (DRL). A major challenge faced by such methods is the limited generalization capability of the active observation policies—specifically, when agents are deployed in environments significantly different from the training environment, their performance in executing AOD tasks may deteriorate substantially \cite{40,41}. Improving the quality of state representation has been proven an effective approach to enhance the generalization ability of agents' active observation policies \cite{42}. Humans and other animals can acquire highly generalizable skills through only a small number of interactions or demonstrations, whereas Machine Learning (ML) systems require massive amounts of samples for learning, and even then, their capabilities are far less reliable than the former. Yann LeCun argues that the underlying reason is that humans and animals can learn internal models of how the world works, namely world models \cite{45}. The core function of world models is to assist AI systems in constructing internal representations of the physical world, based on which prediction, reasoning, and planning can be performed. Meanwhile, world models are typically learned in a task-agnostic and unsupervised manner. Accordingly, this paper proposes a World Model-aided Policy Learning (WMPL) approach. During the policy learning process, a world model is trained synchronously via Self-supervised Learning (SSL) to improve the quality of the agent’s state representations, thereby enhancing the generalization performance of the learned policy.

\begin{figure}[!t]
	\centering
	\includegraphics[width=\linewidth]{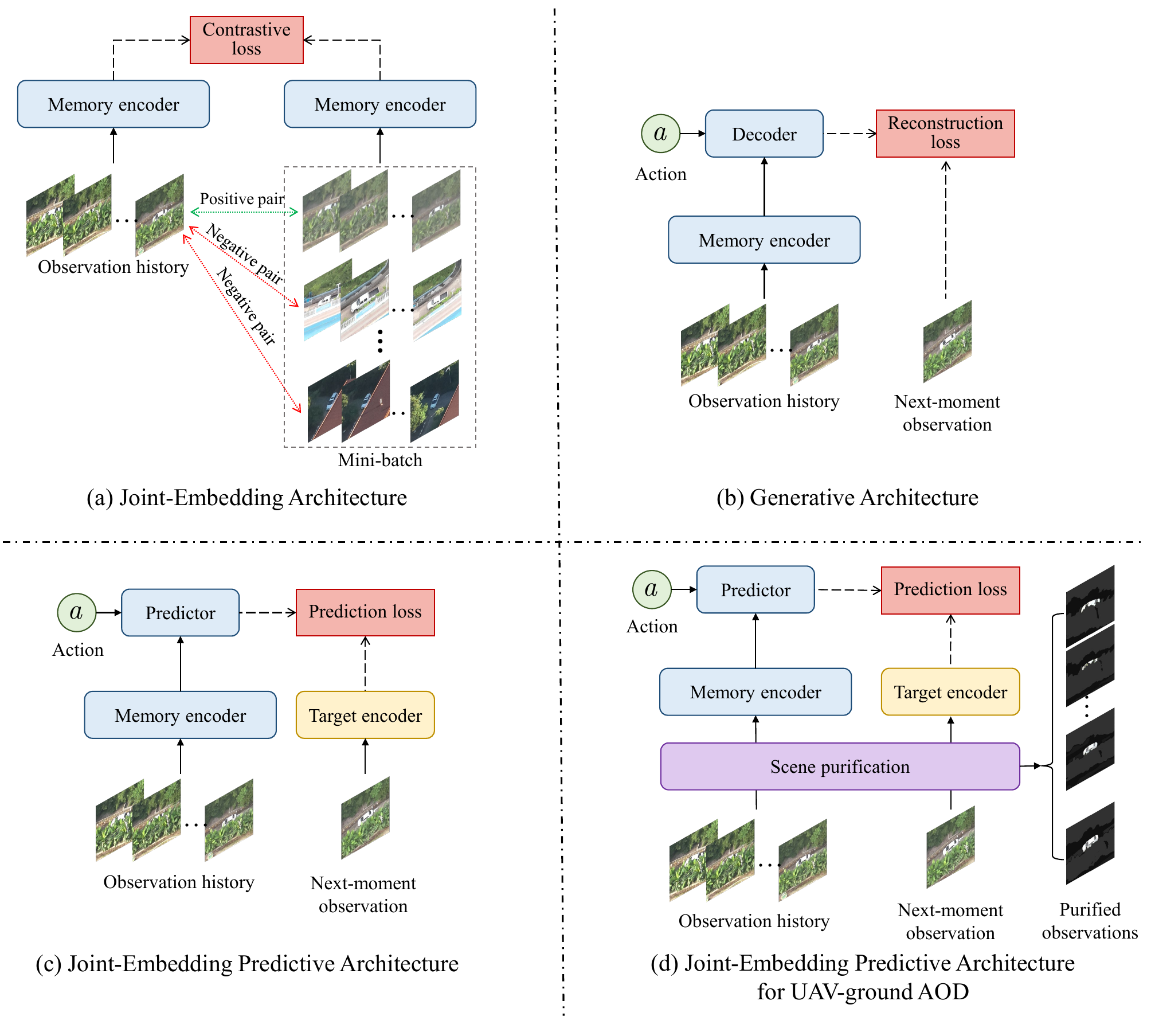}
	\caption{Several categories of SSL architectures applicable to world model learning in embodied perceptual tasks. (a) Joint-Embedding Architecture: Learns to output similar embedding representations for compatible input pairs (x, y) and generate discriminative representations for incompatible input pairs \cite{47}. (b) Generative Architecture: Directly reconstructs the target signal y from the compatible signal x via a decoder network, where the decoder requires an additional conditional variable z to assist the reconstruction process \cite{47}. (c) Joint-Embedding Predictive Architecture: The overall architecture is analogous to the generative counterpart, but the loss function operates in the representation space rather than the original input space \cite{45}. (d) Joint-Embedding Predictive Architecture for UGAOD: Building upon the aforementioned JEPA, prior knowledge related to the AOD task is incorporated into the target encoder to further mitigate the interference of irrelevant information on state representation.}
	\label{fig_2}
\end{figure}

The carousel experiment conducted by Richard Held and Alan Hein demonstrated that a closed ``perception-action" loop is crucial for the maturation of the perceptual system \cite{43}. Inspired by this finding, we argue that for embodied perceptual tasks such as AOD, the causal relationship between an agent's actions and observational information should be modeled within the world model learning framework. Fig.~\ref{fig_2} illustrates several common SSL architectures. Among them, JEPA achieves the causal modeling of action and observation data by performing the next-step observation prediction in the representation space, so it enables the model to learn more task-relevant semantic features for AOD with higher computational efficiency and reduced noise interference \cite{45}. %As shown in Fig.~\ref{fig_2}(a), the Joint-Embedding Architecture (JEA) is presented. For a given anchor sample, SSRL methods adopting this architecture typically construct positive samples through specific data augmentation techniques, while treating all other samples in the same batch as negative samples. Finally, contrastive losses such as InfoNCE \cite{54} are computed to train the model \cite{55,56}. However, the JEA architecture fails to model the action-observation causal relationship and thus cannot assist the agent in understanding the impact of actions on observations. As depicted in Fig.~\ref{fig_2}(b), SSRL methods such as predictive coding employ a Generative Architecture (GA), learning a world model based on the pretext task of next-step observation prediction \cite{57}. Its limitation lies in the inevitable introduction of substantial irrelevant information and noise when reconstructing next-step observations at the pixel level. Additionally, the large volume of reconstructed data and high computational complexity hinder the agent’s state representation from capturing high-level task-relevant semantic features. As illustrated in Fig.~\ref{fig_2}(c), building upon JEA and GA, Yann LeCun proposed the Joint-Embedding Predictive Architecture (JEPA) \cite{45}. JEPA retains the causal modeling of action and observation data from the GA architecture. Furthermore, by performing prediction in the representation space via a non-generative architecture, JEPA enables the model to learn more task-relevant semantic features for AOD with higher computational efficiency, significantly mitigating the interference of task-irrelevant information.

Accordingly, in the proposed WMPL approach, we leverage JEPA for world model training in the UGAOD task for the first time. Meanwhile, we improve JEPA by incorporating task-specific prior knowledge of the AOD task, and propose a Joint-Embedding Predictive Architecture for UAV-ground AOD (AOD-JEPA), as illustrated in Fig.~\ref{fig_2}(d). Guided by human prior knowledge of scene cognition, a high-quality state representation for the AOD task should prioritize task-relevant information (e.g. the target’s appearance, posture, and the spatial positional relationships between the target and its surroundings) while suppressing irrelevant information (e.g., color and texture of the surrounding environment). To this end, we adopt the Segment Anything Model 3 (SAM3) \cite{58} to perform scene purification on observed images: the pixels of the target itself are retained, whereas the color and texture information of surrounding ground objects are filtered out using the segmentation masks generated by SAM3. Moreover, a key advantage of SAM3 is its ability to segment all object instances of the same category in an image based on linguistic prompts. Therefore, we can assign identical grayscale values to the masks of objects belonging to the same category, thereby injecting additional semantic information into the model input. The aforementioned scene purification operation can eliminate the interference of irrelevant information on policy learning and world model training, further improving the quality of the learned state representation and boosting the generalization capability of the active observation policy.

In summary, the main contributions of this paper are threefold:

\begin{enumerate}
	\item[$\bullet$] 
	This paper releases ATRNet-LUDO, the first large-scale real-world dataset for the UGAOD task. Based on this proposed dataset, we can construct a large number of environments suitable for agent interaction. Meanwhile, we establish an evaluation benchmark on the ATRNet-LUDO dataset, defining clear evaluation tasks and metrics to provide an impartial and persuasive experimental platform for this field.
	\item[$\bullet$] 
	We propose the WMPL approach for active observation policy learning, which for the first time leverages JEPA to train a world model for the AOD policy generalization enhancement. Furthermore, by incorporating AOD task-specific prior knowledge into  JEPA, we develop the AOD-JEPA architecture, which further enhances the quality of state representation learning and boosts the generalization capability of the agent’s active observation policy.
	\item[$\bullet$] 
	We conduct extensive comparative experiments on the constructed benchmark, evaluating seven state-of-the-art AOD policy learning methods. Experimental results demonstrate that, compared with traditional passive object perception, the active observation policy improves the target recognition rate by approximately 20 percentage points. In test environments, in contrast to existing policy learning baselines, the WMPL method achieves a 2–3 percentage point gain in target recognition rate while maintaining nearly unchanged UAV motion costs. 
\end{enumerate} 

The remainder of this paper is organized as follows. Sec.~\ref{Related Work} briefly reviews the datasets and methods relevant to the AOD task. Sec.~\ref{Dataset} presents the collection and construction process as well as the statistical analysis of the ATRNet-LUDO dataset. In Sec.~\ref{Approach}, we elaborate on the proposed WMPL approach for active observation policy learning. Sec.~\ref{Benchmark} introduces the evaluation benchmark constructed on the basis of the ATRNet-LUDO dataset. Sec.~\ref{Experiments} includes the comparative experiments conducted on the evaluation benchmark and the corresponding result analysis. Finally, Sec.~\ref{Conclusion} summarizes the entire paper and provides prospects for future research directions.

\section{Related Work}
\label{Related Work}

\begin{table*}[tb]
	\centering
	\caption{Comparison of ATRNet-LUDO with Related UAV Image Object Detection Datasets and Active Object Detection Datasets}
	\label{table_datasets}
	\renewcommand\arraystretch{1.3}
	\resizebox{\linewidth}{!}{%
		%\begin{tabular}{lcp{1.7cm}<{\Centering} p{2.4cm}<{\Centering} p{1.3cm}<{\Centering} p{1.7cm}<{\Centering} p{6.5cm}<{\RaggedRight}} 
		\begin{tabular}{lccccccccc} 
			\toprule
			\multicolumn{1}{c}{\multirow{1}{*}{\textbf{Dataset}}} & \multirow{1}{*}{\textbf{Task}} & \multirow{1}{*}{\textbf{Scenario}} & \textbf{Measured} & \textbf{Multi-view} & \textbf{Modality} & \multicolumn{1}{c}{\multirow{1}{*}{\textbf{Target number}}} & \textbf{Target type} & \textbf{Environment number} & \textbf{Size}\\ 
			\cmidrule(lr){1-10}
			VisDrone~\cite{33} & Object detection & Outdoor, UAV & 	
			\checkmark &  \xmark & RGB & 10 & Vehicles, pedestrian & - & 10209\\
			DroneVehicle~\cite{34} & Object detection & Outdoor, UAV & \checkmark & \xmark & RGB, Infrared & 5 & Vehicles & - & 28439\\
			UAVDT~\cite{35} & Object detection & Outdoor, UAV & \checkmark & \xmark & RGB & 3 & Vehicles & - & 80000\\
			AU-AIR~\cite{36} & Object detection & Outdoor, UAV & \checkmark & \xmark & RGB & 8 & Vehicles, pedestrian & - & 32823\\
			\cmidrule(lr){1-10}
			AVD~\cite{28} & AOD & Indoor, robot & \checkmark & \checkmark & RGB-D & 33 & Daily necessities & 17 & 20916\\
			T-LESS~\cite{37} & AOD & Indoor, robot & \checkmark & \checkmark & RGB & 30 & Industrial components & 20 & 10000+\\
			R3ED~\cite{38} & AOD & Indoor, robot & \checkmark & \checkmark & Point cloud & 12 & Furniture, home appliances & 7 & 5800\\
			UEVAVD~\cite{39} & AOD & Outdoor, UAV & \xmark & \checkmark & RGB & 5 & Vehicles & 150 & 90750\\
			CARLA-AOD~\cite{21} & AOD & Outdoor, UAV & \xmark & \checkmark & RGB & 4 & Vehicles & 18 & 2160\\
			ATRNet-LUDO & AOD & Outdoor, UAV & \checkmark & \checkmark & RGB & 10 & Vehicles & 2000 & 1210000\\
			\bottomrule
	\end{tabular}}
\end{table*}

\subsection{Active Object Detection}
AOD differs from conventional static object detection and Target-Driven Visual Navigation (TDVN). Traditional detection pipelines localize and categorize targets from fixed single-view inputs without closing the loop between data collection and observation quality optimization. In contrast, AOD leverages embodied agents' motion planning to collect informative multi-view observations.

TDVN, also named Active Object Search (AOS), was first proposed in \cite{27}. Its core task is to navigate an agent toward a predefined target that is invisible in initial frames and must become observable after movement. Although TDVN and AOD both require visual semantic extraction and autonomous path planning, they differ fundamentally in task objectives. TDVN pursues unseen predefined targets, while AOD validates the identity of suspicious targets already visible with fixed positions in initial observations.

DRL has become the dominant solution for AOD by casting the task as sequential decision making. Ammirato et al. pioneered REINFORCE-based indoor AOD and released the Active Vision Dataset (AVD) for standardized algorithm evaluation \cite{28}. Han et al. presented Multistep Action Prediction (MAP) built upon dueling DQN with prioritized experience replay to jointly predict action types and magnitudes \cite{29,30}. Fang et al. further extended MAP to SSL-MAP, which adopts multi-view self-supervised representation learning to boost DRL sample efficiency \cite{13}. Subsequent works improve state encoding and policy fine-tuning. Liu et al. integrated cropped target patches into state representation and designed dedicated reward functions for smooth target approaching \cite{31}. Ding et al. developed a Decision Transformer online fine-tuning framework: it initializes policies via offline expert datasets and refines policies through real environmental exploration \cite{32}.

\subsection{Relevant Datasets for Active Object Detection}
Task-specific datasets greatly promote AOD research: they enable low-cost, risk-free simulation of agent-environment interactions and provide standardized evaluation benchmarks \cite{28}. Table~\ref{table_datasets} lists representative indoor AOD datasets AVD \cite{28}, T-LESS \cite{37} and R3ED \cite{38}, all collected via mobile robots with dense indoor multi-view sampling. AVD is the first real-world indoor AOD dataset with 17 scenes, 20,916 RGB-D frames and 33 annotated target classes. R3ED contains 5,800 multi-view point cloud samples across 7 indoor scenes with 12 categories, while T-LESS offers multi-view industrial object data captured by adjusting camera angles and turntable rotation. Though supporting trajectory simulation for AOD analysis, all these datasets are limited to indoor scenes and lack outdoor UAV-AOD support.

Conventional UAV detection benchmarks (VisDrone \cite{33}, DroneVehicle \cite{34}, UAVDT \cite{35}, AU-AIR \cite{12}) collect aerial pedestrian and vehicle images but only contain sparse viewpoints per scene, incapable of simulating continuous flight observation for UGAOD. Our group built the UEVAVD dataset \cite{39} in 2024 via Unreal Engine and AirSim with 150 UGAOD simulation environments to mitigate this issue. Another synthetic benchmark CARLA-AOD \cite{44} adopts hemispherical sampling to gather panoramic data of 24 viewpoints and 5 scales over 18 terrain types \cite{21}. Nevertheless, both simulation datasets suffer from synthetic-real domain gaps and limited diversity of targets and scenes, weakening their capacity for fair comparison of AOD policy algorithms. To overcome such limitations, we build a new dataset covering 40 regions and 10 target classes, consisting of 121,000 panoramic images and 1.21 million local target patches, which supports 200 multi-target and 2,000 single-target AOD simulation environments.

\subsection{Joint Embedding Predictive Architecture}
Biological intelligence achieves high learning efficiency via innate world models for physical reasoning, prediction and planning, while standard machine learning systems lack such capabilities \cite{45}. Learning world models through self-supervised learning (SSL) is therefore a core challenge for modern artificial intelligence. SSL aims to model inherent correlations among raw input data.

Built on Joint-Embedding Architectures (JEA) and generative models, JEPA optimizes predictive loss in latent representation space instead of pixel space \cite{45}. This design avoids reconstructing trivial pixel-level details, automatically filters irrelevant background features, and significantly reduces computational overhead. Unlike JEA, JEPA does not enforce invariance toward handcrafted data augmentations; instead, it learns mutually predictable cross-view representations conditioned on auxiliary scene information $z$.

JEPA has achieved promising performance across extensive tasks, including image classification \cite{47}, video understanding \cite{48}, skeleton action recognition \cite{49}, wireless channel estimation \cite{50}, SAR target recognition \cite{51,52}, and audio classification \cite{53}. These applications validate the effectiveness of JEPA's core idea: treating latent feature prediction as the self-supervised objective facilitates extracting high-quality multi-modal representations. This work makes the first attempt to adopt JEPA for learning active observation policies in AOD. Tailored to the embodied active detection task, we further propose AOD-JEPA based on the original JEPA backbone to enhance agent state representation quality.

\section{Dataset Collection and Construction}
\label{Dataset}
\subsection{Target and Scene Configuration}

\begin{figure}[!t]
	\centering
	\includegraphics[width=\columnwidth]{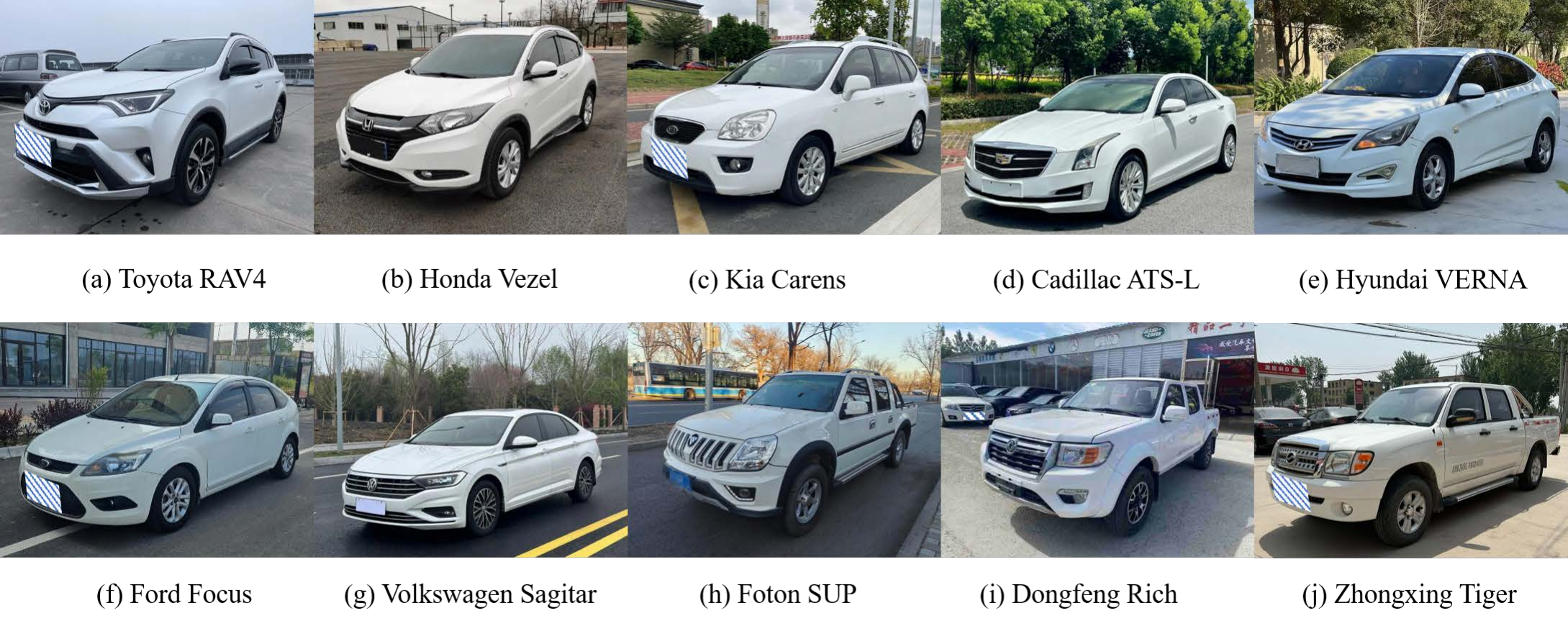}
	\caption{Showcase of the ten vehicle target types used in the ATRNet-LUDO dataset. The first three are SUVs, the middle four are sedans, and the last three belong to pickup trucks. All vehicles share the white color, with differences lying in their exterior shapes and component structures.}
	\label{fig_3}
\end{figure}

Vehicles constitute dominant targets in UAV imagery, and UAV-based vehicle detection is of great practical significance. Accordingly, this dataset takes civilian small vehicles as its primary research objects, covering ten common models: Toyota RAV4, Honda Vezel, Kia Carens, Cadillac ATS-L, Hyundai VERNA, Ford Focus, Volkswagen Sagitar, Foton SUP, Dongfeng Rich, and Zhongxing Tiger (Fig.~\ref{fig_3}). The selected vehicles include three SUVs, four sedans, and three pickup trucks, requiring UAV perception systems to achieve fine-grained category recognition. To avoid simplistic color-based classification and focus on discriminative structural features, all vehicle targets in the dataset are uniformly white. This design eliminates color interference, making intra-class and inter-class differences mainly reflected in geometric contours and component structures, which poses a more realistic and challenging fine-grained recognition task.

\begin{figure}[!t]
	\centering
	\includegraphics[width=\columnwidth]{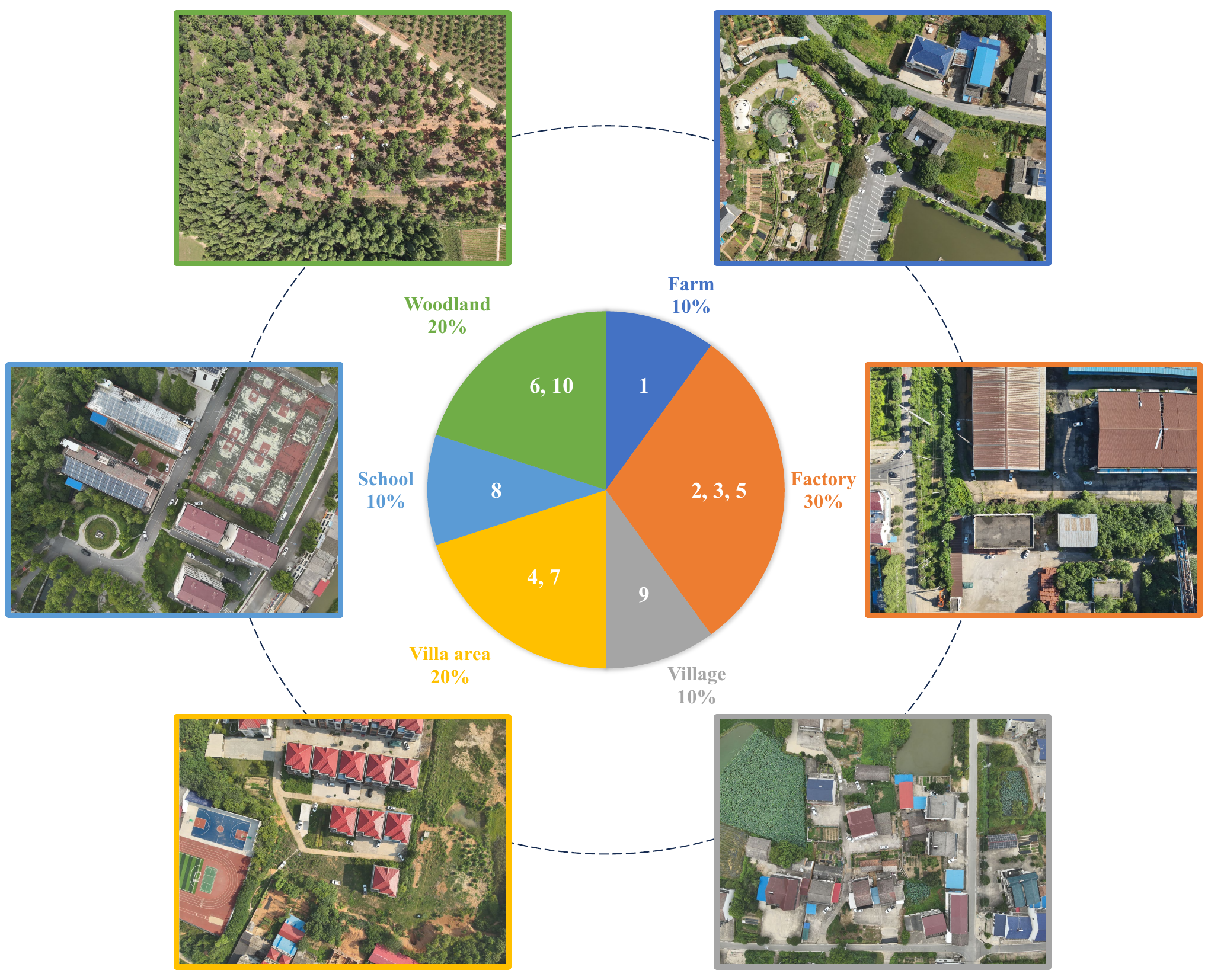}
	\caption{Showcase of scenes and corresponding data in the ATRNet-LUDO dataset. Occlusion of vehicle targets is common in the selected scenarios, including occlusions caused by trees and buildings.}
	\label{fig_4}
\end{figure}

On the other hand, our dataset focuses on AOD under various occluded scenarios. Considering that the primary ground objects prone to occluding vehicle targets are trees and buildings, we selected ten regions, covering six typical occlusion-prone scenarios including factories, campuses, villa districts, villages, farms, and woodlands. Example images corresponding to each scenario are presented in Fig.\ref{fig_4}. Meanwhile, we statistically analyzed the proportion of data collected across different scenarios, among which UAV images captured in factories, woodlands, and villa districts account for a larger share. The numbers in the pie chart represent the corresponding regional serial numbers in the dataset. Furthermore, to acquire data for training the perception module, we additionally designated an open area to place the ten targets and collected multi-view imaging results under occlusion-free conditions.

\subsection{Data Collection}
\begin{figure}[!t]
	\centering
	\includegraphics[width=\columnwidth]{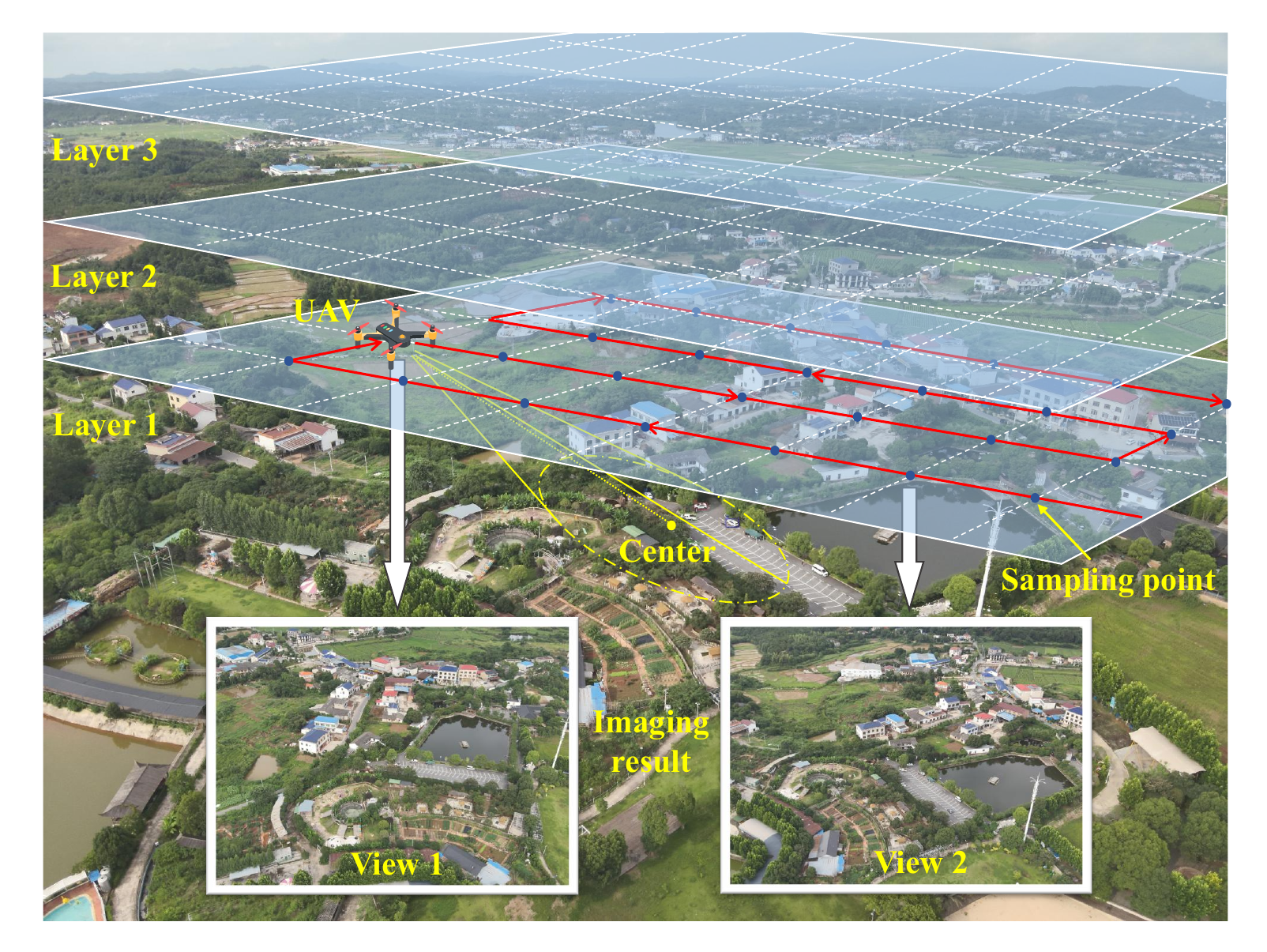}
	\caption{Schematic diagram of UAV-based dense multi-view sampling for ground targets. The UAV traverses the predefined sampling points across all layers along the illustrated trajectory above the selected region, performing multi-view imaging of vehicle targets near the center of the sub-region. The predefined sampling points are densely arranged as the vertices of square grids.}
	\label{fig_6}
\end{figure}

\begin{figure*}[!t]
	\centering
	\includegraphics[width=0.8\linewidth]{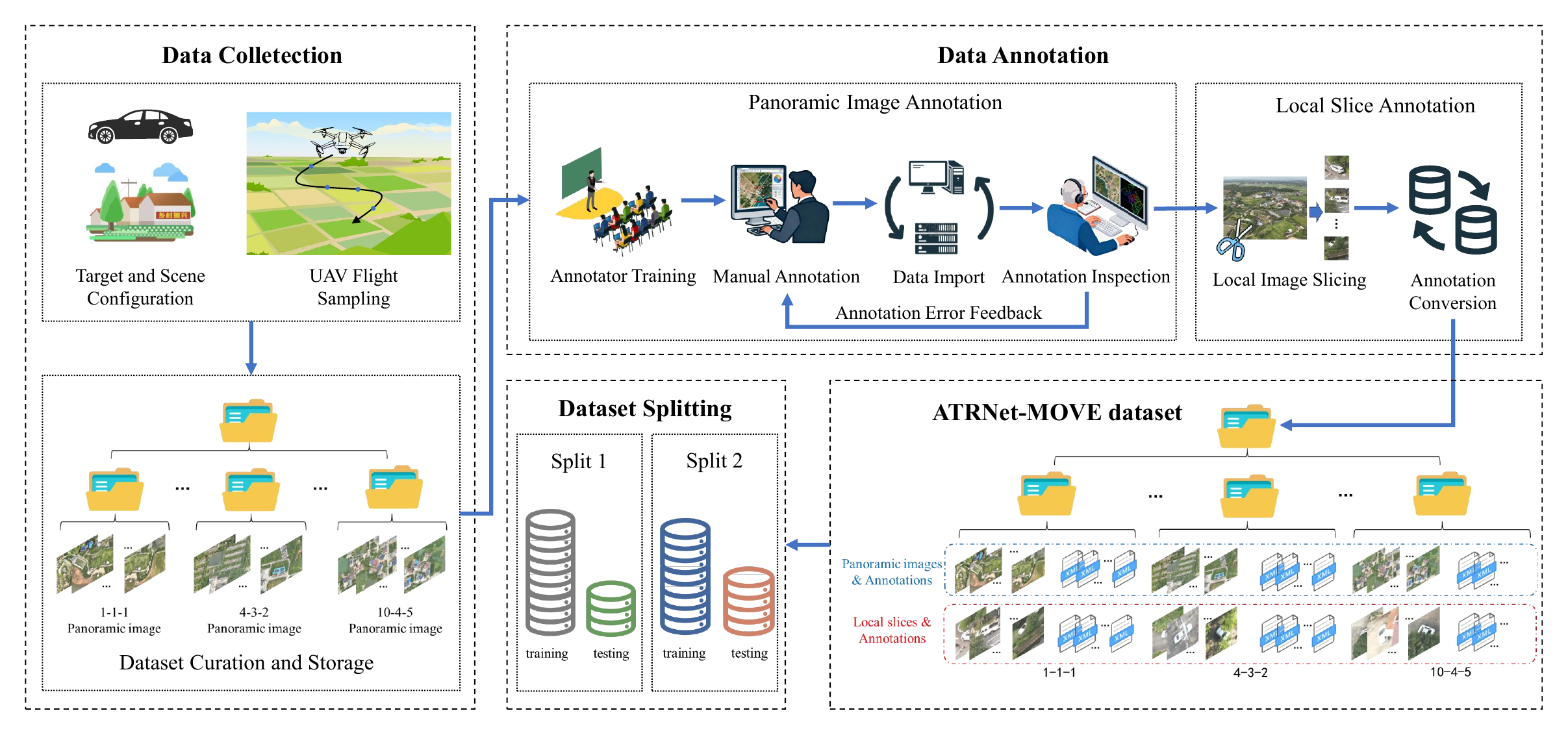}
	\caption{Flowchart of the ATRNet-LUDO Dataset Construction. Based on the target and scenario configurations, a UAV is deployed to perform multi-view imaging and sampling of ground vehicle targets across multiple scenarios according to the acquisition protocol. The collected data are organized and stored to form a preliminary dataset. Subsequently, annotators label and inspect the panoramic images; on this basis, local image slicing and annotation conversion are conducted. Finally, the dataset is organized by environment IDs to form the finalized ATRNet-LUDO dataset, which is then partitioned for model training and evaluation.}
	\label{fig_5}
\end{figure*}

The ATRNet-LUDO dataset is collected via UAV aerial sampling at predefined spatial positions over multiple ground regions. Its diversity is systematically enriched in terms of scene coverage, target categories, and spatial target–background relationships. Specifically, each investigated region is divided into four 80 m $\times$ 80 m subregions, where targets are randomly oriented and placed near occluders to simulate complex real-world distributions. To ensure sufficient sampling diversity, target layouts are reset after each aerial capture, yielding five repeated sampling trials for every subregion. Each trial corresponds to an independent interactive environment with a unique ID. Based on this configurable sampling strategy, multi-view data are acquired for ten target categories across 40 subregions, producing 121,000 panoramic images and 1,210,000 local target patches. The constructed dataset supports 200 multi-target and 2000 single-target AOD simulation environments. All panoramic images are captured at a resolution of 8000$\times$6000 pixels, while each local patch is cropped as a 300$\times$300 pixel region centered on the target, preserving fine-grained target and surrounding contextual details.

Fig.~\ref{fig_6} visualizes UAV multi-view sampling for 10 vehicle classes within a single subregion. After target placement around the subregion center, a DJI Mavic 3 Enterprise UAV conducts aerial capture; its onboard camera delivers a ground sampling distance of 0.036 m at 100 m altitude. Sampling points are organized into five altitude layers with identical horizontal coordinates, starting at $h_\text{min}=120\,\text{m}$ and spaced vertically by $\Delta h=30\,\text{m}$. Each layer features a $10\times10$ grid aligned with latitude/longitude lines, containing 121 vertices as sampling nodes separated by $\Delta l=40\,\text{m}$. In total, one subregion covers $5\times121=605$ sampling positions. Following the red flight trajectory in Fig.~\ref{fig_6}, the UAV captures imagery at all nodes while keeping the camera oriented toward the subregion center. This yields 605 multi-view panoramic frames and 6,050 cropped local target patches per subregion.

\subsection{Data Annotation}
\begin{figure*}[!t]
	\centering
	\includegraphics[width=0.8\linewidth]{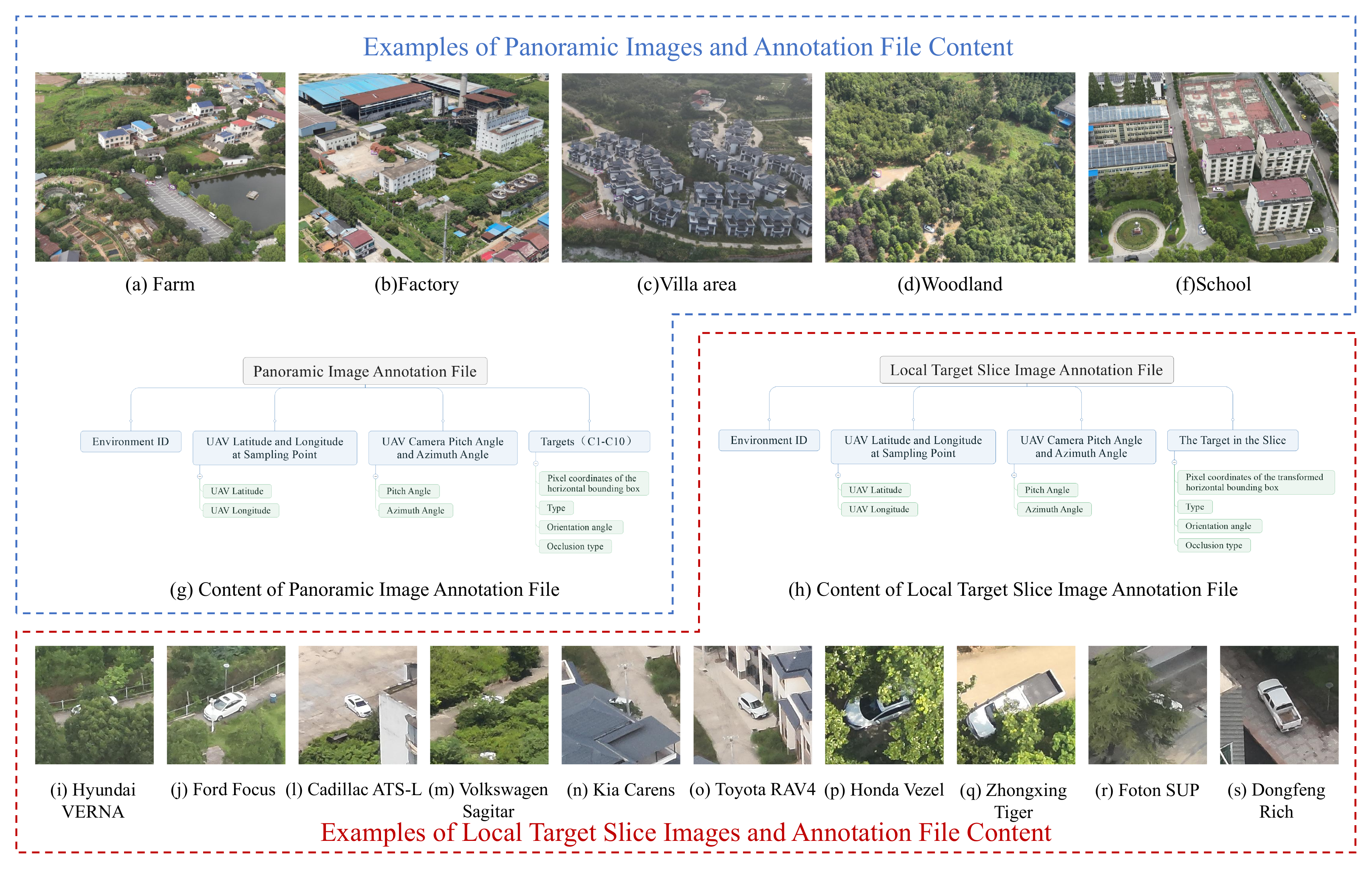}
	\caption{Visualization of panoramic aerial images, local target patches and annotation files from the ATRNet-LUDO dataset.}
	\label{fig_7}
\end{figure*}

\begin{figure*}[!t]
	\centering

	\includegraphics[width=\linewidth]{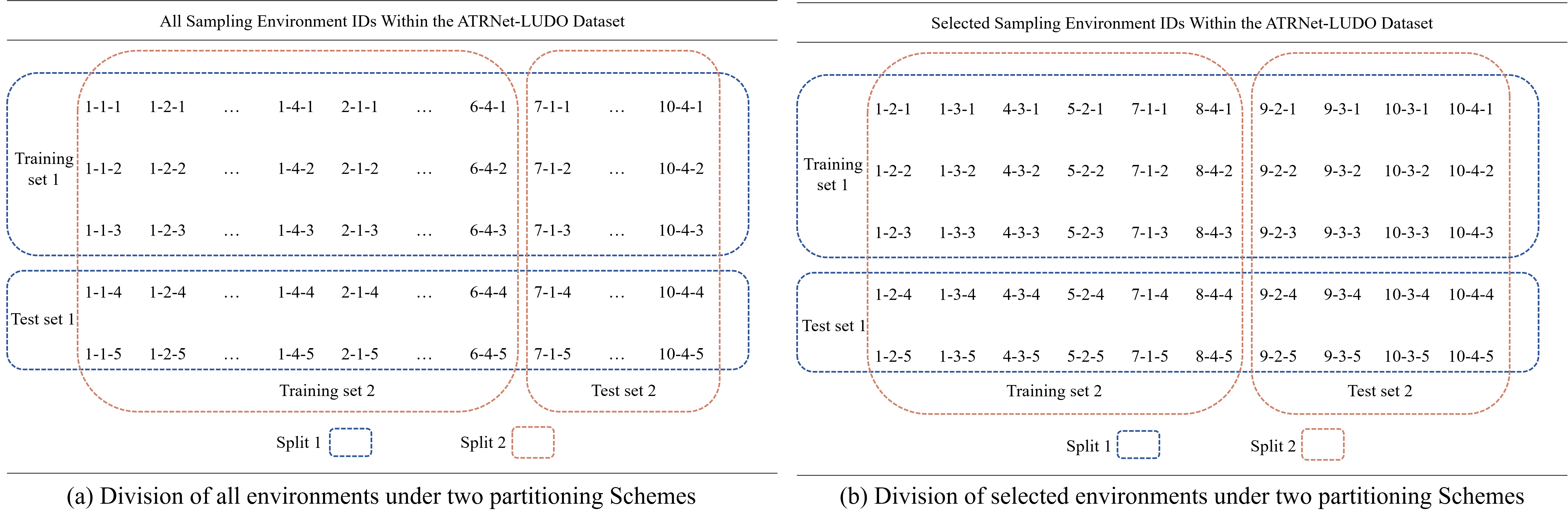}
	\caption{Splits and statistics of the ATRNet-LUDO's training and test environments.}
	\label{fig_split}
\end{figure*}

\begin{figure*}[!ht]
	\centering
	\includegraphics[width=0.9\linewidth]{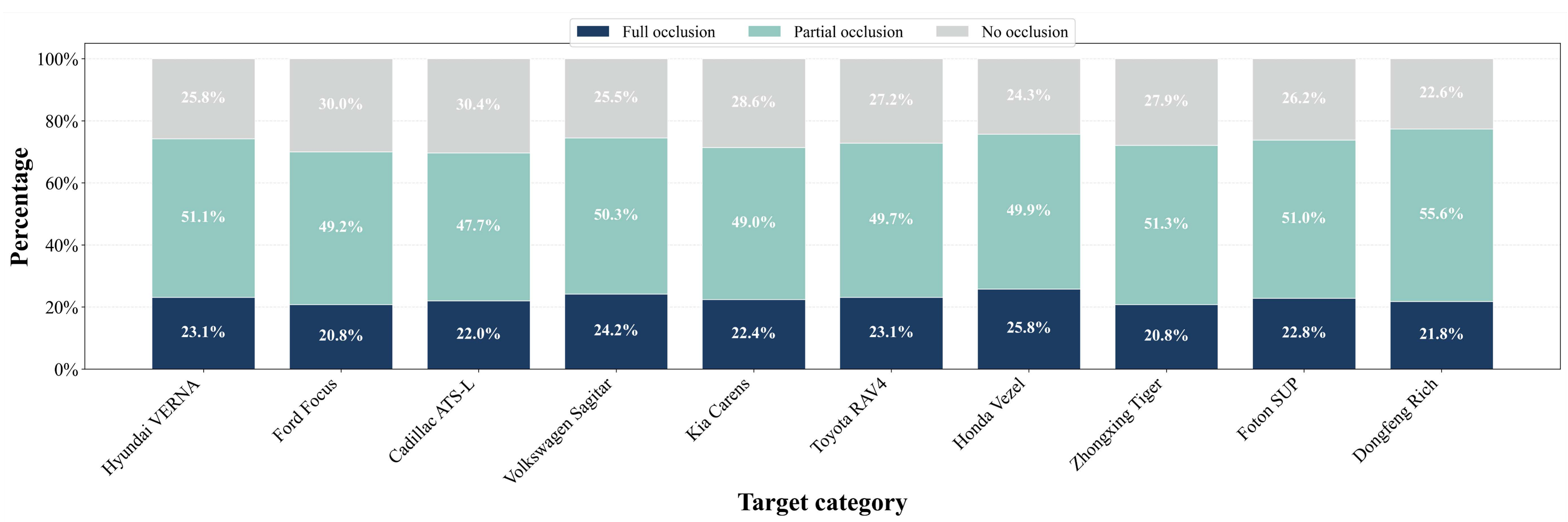}
	\caption{Occlusion statistics and slice example demonstration for the ten vehicle target categories in the ATRNet-LUDO dataset.}
	\label{fig_9}
\end{figure*}

As shown in Fig.~\ref{fig_5}, all collected panoramic images are sorted by environment ID before annotation. The annotation pipeline consists of panoramic image labeling and local patch labeling. Approximately thirty annotators were professionally trained on the X-AnyLabeling tool \cite{X-AnyLabeling} and certified via qualification tests prior to official labeling. In panoramic annotation, each target is bounded with horizontal boxes and labeled with category and occlusion attributes. Meanwhile, auxiliary metadata, including environment ID, UAV geolocation, camera pitch/azimuth, and target orientation, is synchronously recorded. To guarantee label quality, we adopt random manual inspection and algorithmic validation. Annotations with missing targets, incorrect categories, inaccurate bounding boxes, or redundant information are returned for revision.

The second annotation phase involves local slicing and label conversion. Target-centered 300$\times$300 local patches are cropped from panoramic images. For partially occluded targets, patch centers are directly derived from annotated bounding-box centers, while fully occluded targets require coordinate prediction. We build a deep neural network to infer their pixel positions using scene target geolocations, UAV poses, and subregion center coordinates, yielding an average prediction error of 53 pixels. Most annotation attributes, including environment ID, UAV geolocation, camera angles, target orientation, category, and occlusion status, are inherited from panoramic labels. Only bounding-box coordinates are transformed from panoramic to patch coordinate systems. For fully occluded targets, bounding boxes are uniformly set to $[99, 99, 201, 201]$. All annotations are stored in XML format. Representative panoramic images, local patches, and annotation examples are visualized in Fig.~\ref{fig_7}.

\subsection{Dataset Splitting}
\label{subsec3.4}
To construct a unified benchmark for UGAOD policy evaluation, we partition the ATRNet-LUDO dataset to test policy generalization under two realistic environmental discrepancies: varying target-background layouts within shared subregions, and completely unseen sampling regions. Correspondingly, two data-splitting strategies are designed to evaluate policy generalization at different difficulty levels. As shown in Fig.~\ref{fig_split}(a), the original 200 multi-target environments (2000 single-target environments) are divided via two schemes. The first splits data by sampling trials, using the first three trials for training and the rest for testing. The second splits data by geographic region, adopting the first six regions for training and the remaining four for testing.

Considering the massive trajectory diversity and high redundancy of the original environments, redundant backgrounds and similar target layouts lead to heavy computational costs and slow policy convergence. Therefore, we select 50 representative multi-target environments (500 single-target environments) while maintaining scene diversity, as visualized in Fig.~\ref{fig_split}(b). The two aforementioned partitioning schemes are reapplied to the refined subset to accelerate evaluation without degrading generalization assessment.
\subsection{Statistics and Analysis}

A core advantage of ATRNet-LUDO lies in abundant target-background spatial layouts generating diverse occlusion patterns, supporting training and evaluation of active perception algorithms under occlusions. Fig.~\ref{fig_9} plots occlusion statistics across all target classes: more than 70\% of targets suffer occlusion, dominated by partial occlusion alongside fully occluded samples.

\begin{figure}[ht]
	\centering
	\includegraphics[width=0.8\columnwidth]{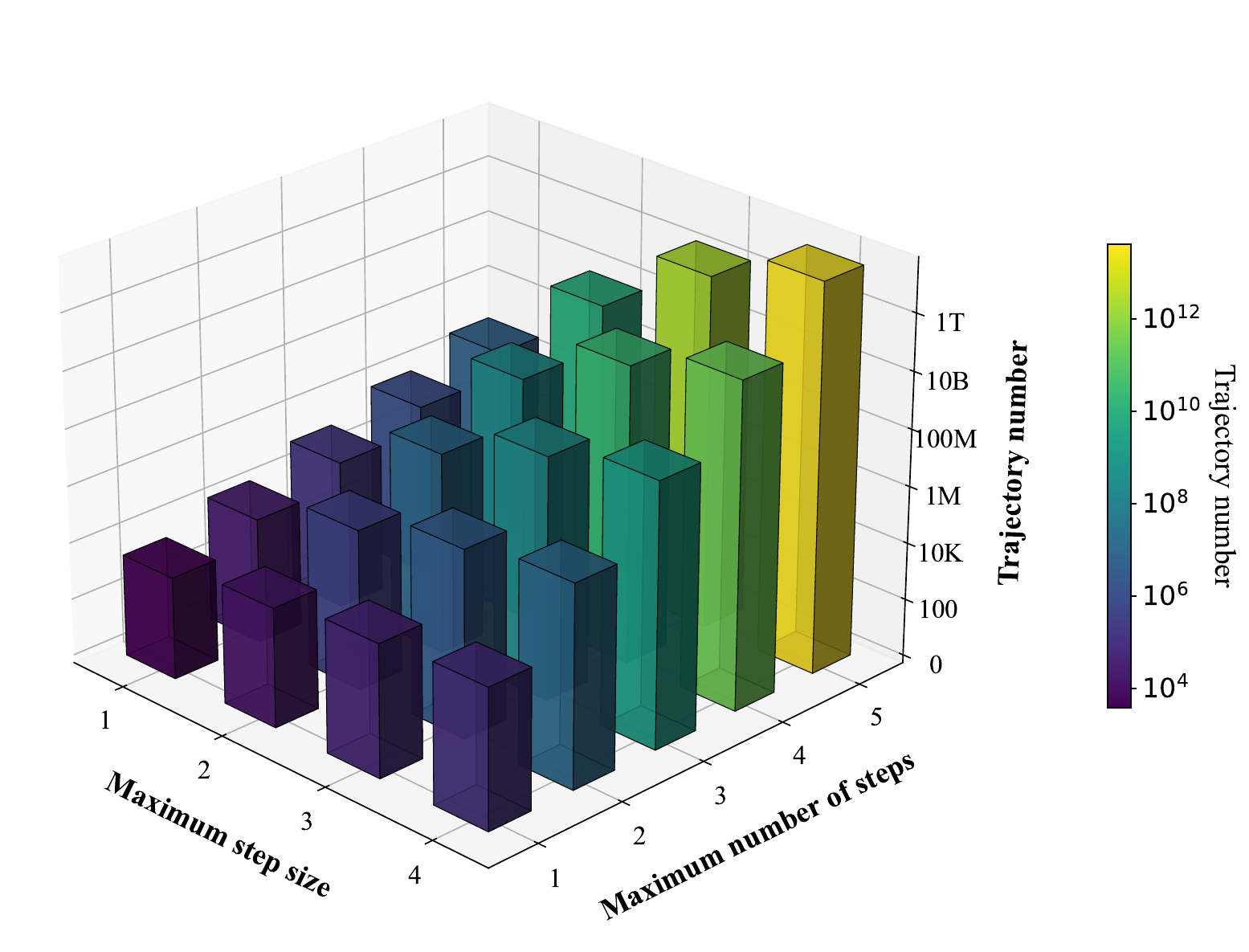}
	\caption{Statistical plot of simulated trajectory count under different maximum step sizes and maximum step numbers in a single environment. The trajectory count grows exponentially with increasing maximum step number and maximum step size. When both of them are set to 3, the number of simulated UAV flight trajectories can easily exceed 10 million.}
	\label{fig_8}
\end{figure}

In this section, a comprehensive statistical analysis of the ATRNET-LUDO dataset is conducted to help readers better understand its characteristics. Adjacent multi-view images simulate UAV flight trajectories for active policy training and benchmarking. Fig.~\ref{fig_8} counts feasible trajectories under varying maximum step size $A$ and maximum step count $T_\text{max}$. At each step, the UAV selects a movement of $1$–$A$ units or terminates, repeating for up to $T_\text{max}$ steps. Trajectory quantity rises exponentially with $A$ and $T_\text{max}$; with $A=T_\text{max}=3$, the number surpasses 10 million, demonstrating the dataset’s capacity to cover diverse aerial observation workflows.

\begin{figure}[!t]
	\centering
	\includegraphics[width=\columnwidth]{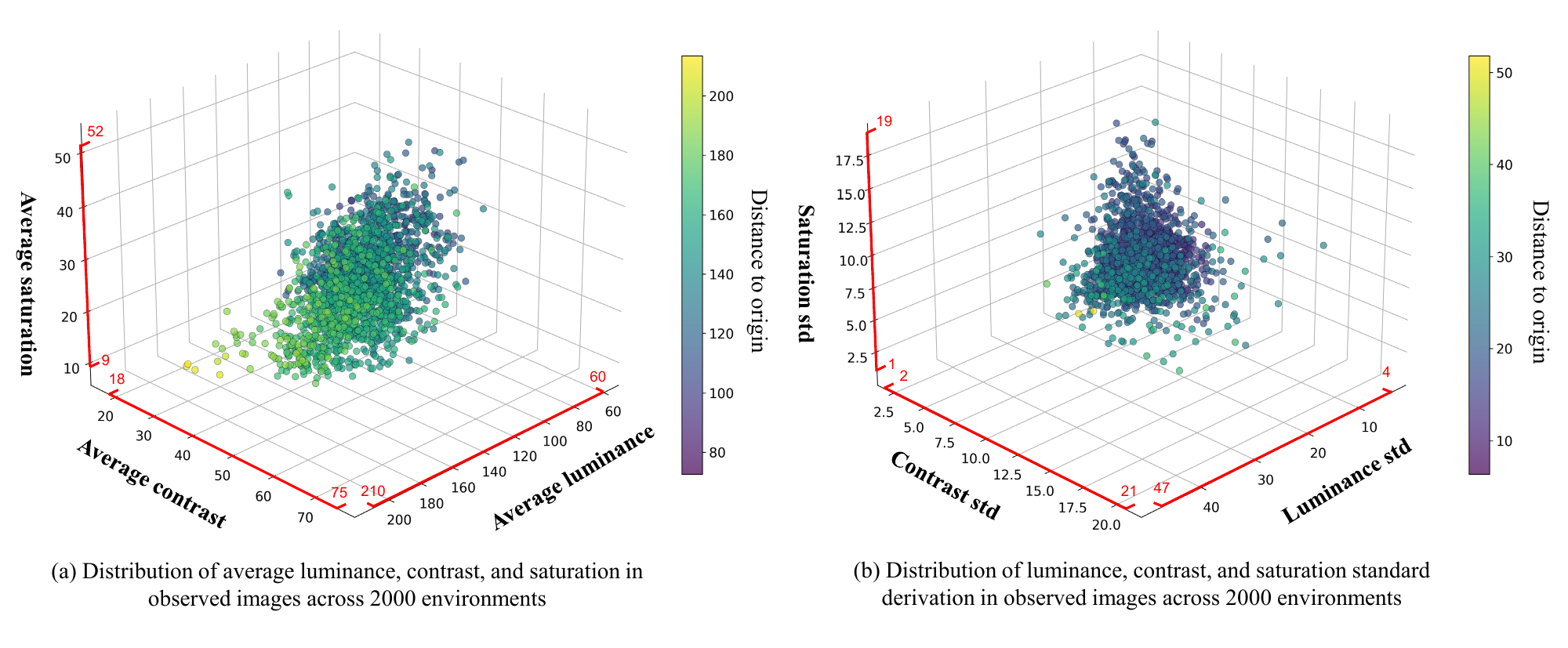}
	\caption{Distribution of statistical values of observation image attributes across 2000 environments in ATRNet-LUDO.}
	\label{fig_10}
\end{figure}

We statistically analyze the attribute diversity of the 1,210,000 local patches from 2000 environments to verify the dataset’s richness (Fig.~\ref{fig_10}). For each patch, luminance is quantified by the average grayscale pixel value, contrast by the grayscale standard deviation, and saturation by the standard deviation of the HSV saturation channel. In Fig.~\ref{fig_10}(a), each point denotes the averaged brightness, contrast, and saturation of the 605 multi-view patches per environment, while Fig.~\ref{fig_10}(b) corresponds to their attribute standard deviations. Data collection spanning five months and daily full daylight hours yields diverse illumination, fog, and weather conditions. This leads to wide distribution coverage across all three attribute dimensions (highlighted in red), as shown in Fig.~\ref{fig_10}(a). Meanwhile, viewpoint variations induce observable attribute fluctuations within individual environments (Fig.~\ref{fig_10}(b)). These results demonstrate that ATRNet-LUDO covers abundant, realistic imaging variations, providing high-quality data for UAV–ground active detection and recognition under complex environmental conditions.

Furthermore, the statistical results indicate that the spatial resolution at the center of the panoramic images ranges approximately from 0.04 m/pixel to 0.13 m/pixel, with both the average and median resolutions being 0.09 m/pixel, demonstrating that the dataset images overall possess high resolution.

\section{Approach}
\label{Approach}
This section first formulates the AOD problem and reviews how to address this problem within the DRL framework. Next, it introduces the proposed WMPL approach for active observation policy learning, elaborating on the joint training mechanism of the world model and the agent’s policy network in the WMPL method.

\subsection{Problem Formulation}
\begin{figure}[!t]
	\centering
	\includegraphics[width=\columnwidth]{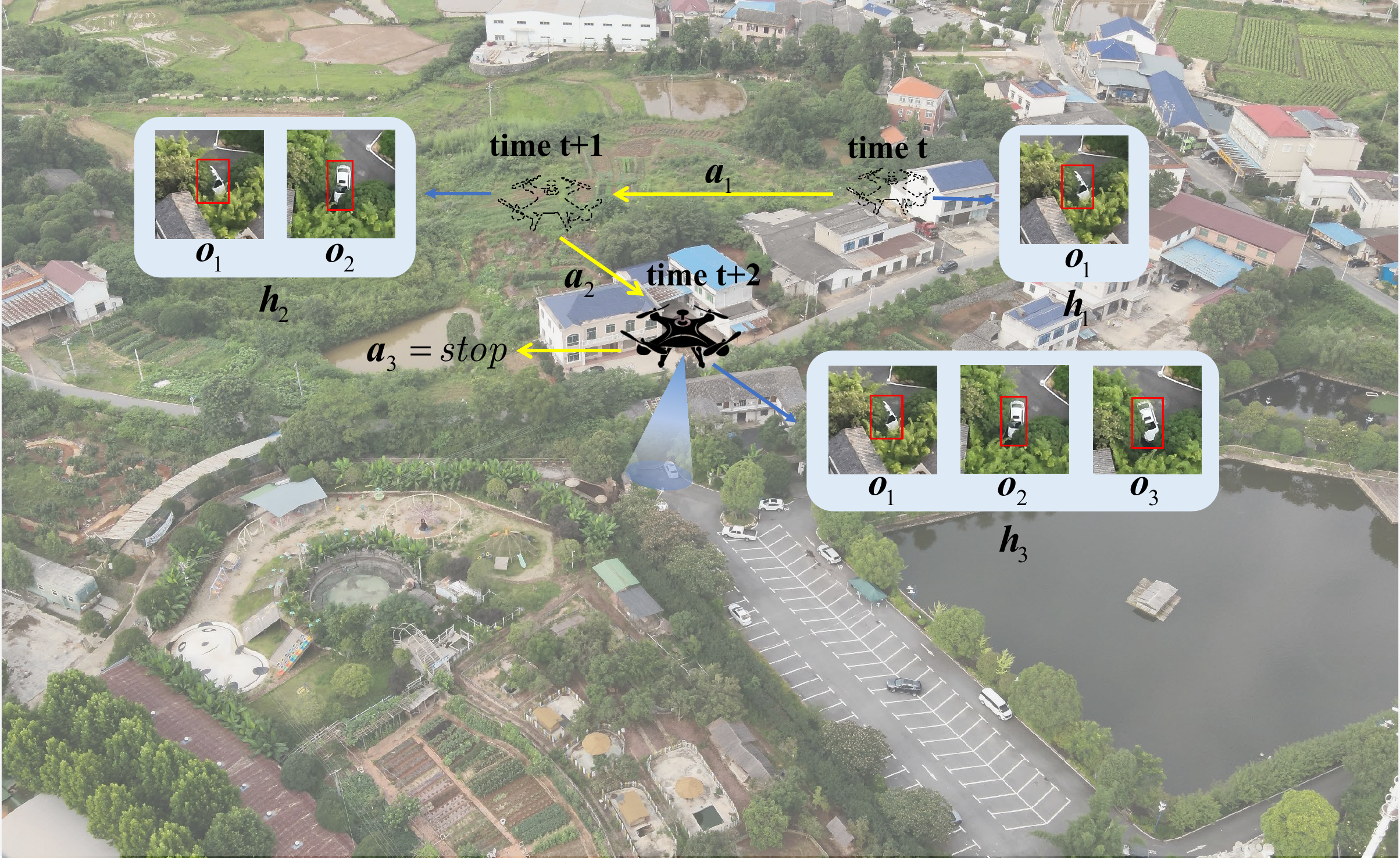}
	\caption{Schematic diagram of the UAV-ground active object detection scenario. At each time step, the policy model of the UAV agent makes decisions based on historical observation information to guide itself toward the optimal viewing angle. Ideally, when the policy model determines that the current observations are sufficient for target recognition, it promptly issues a stop command to terminate the UAV's movement.}
	\label{fig_11}
\end{figure}

The UGAOD scenario is illustrated in Fig.~\ref{fig_11}. When a suspicious target is detected by the UAV at the initial moment, the AOD process is initiated. The target observation image at this moment is denoted as $\bm{o}_1$. At each subsequent time step, the agent's decision-making module selects an action $\bm{a}$ based on the observation history $\bm{h}$, which includes all observation images and executed actions from time step $t$ onwards. The action $\bm{a}$ can either guide the UAV to move toward the optimal observation viewpoint or terminate the AOD process. The objective of the AOD task is to enable the UAV to perform path planning in the environment using historical observation data, aiming to move to the optimal observation position with minimal time cost and improve the accuracy of target recognition. Therefore, we formulate the problem of learning an active observation policy for the agent as the following optimization problem:

\begin{equation}
	\label{eq1}
	\mathop {\min }\limits_\omega  {\mathbb{E}_{({\bm{o}_1},y)\sim{\cal D}}}[{{\cal L}_c}(P({F_\text{env}}({\bm{o}_1},{\pi _\omega })),y) + \lambda L({\bm{o}_1},{\pi _\omega })].
\end{equation}

Here, $\bm{o}$ denotes the agent's observation data for the region of interest, consisting of three components: the observation image $\bm{I}$, the target bounding box $\bm{b}$, and the UAV lens orientation (angle relative to the north direction) $\bm{\alpha}$. $y$ represents the true identity label of the target, and $\mathcal{D}$ denotes the data distribution of $(\bm{o}_1, y)$. $\pi_\omega$ with parameter $\omega$ is the policy model used for decision-making; $\pi_\omega$ selects actions based on the historical observation data $\bm{h}_t = (\bm{o}_1, \bm{a}_1, \bm{o}_2, \bm{a}_2, \dots, \bm{o}_t)$, i.e., $\bm{a}_t = \pi_\omega(\bm{h}_t)$. Subsequently, the motion control module guides the agent to move according to the selected action. At this point, the environment feeds back new observation data $\bm{o}_{t+1}$ to the agent, which then enters the next "decision-motion-observation" cycle. It is generally assumed that the target position $\bm{b}_{t+1}$ in $\bm{o}_{t+1}$ can be easily determined using object tracking algorithms \cite{30}. The termination timing of the AOD process is also decided by $\pi_\omega$. The function $F_{\text{env}}(\cdot)$ is determined by the environment where the agent is located: given $\bm{o}_1$ and $\pi_\omega$ as inputs, the environment returns the optimal observation $\bm{o}^*$ finally obtained by the agent under the policy $\pi_\omega$, namely,

\begin{equation}
	\label{eq2}
	F_{\text{env}}(\bm{o}_1,\pi_\omega)=\bm{o}^*.
\end{equation}

During this mapping process, the decision module can perform multiple decisions, and the agent interacts with the environment multiple times. $P(\cdot)$ denotes the agent's perception model, which can determine the target category based on the observation data. $\mathcal{L}_c$ represents the cross-entropy loss function. On the other hand, we incorporate the total path length throughout the motion process into the optimization objective, which is calculated by the function $L(\bm{o}_1, \pi_\omega)$. The proportion of the two components in the optimization objective is adjusted by the coefficient $\lambda$.

There exists a highly complex nonlinear mapping relationship between high-dimensional image inputs and optimal action selection. Moreover, understanding the 3D world based on a small number of 2D observation images introduces additional uncertainties into this complex mapping. Therefore, traditional control techniques relying on accurate system models are not suitable for the AOD problem, and a data-driven approach should be adopted instead. Considering that AOD is essentially a sequential decision-making problem based on high-dimensional observations, existing methods generally address it within the DRL framework. First, the AOD problem is modeled as a Partially Observable Markov Decision Process (POMDP), which can be described by a 7-tuple $\left\langle \mathcal{S}, \mathcal{A}, \mathcal{O}, T, \Omega, R, \gamma \right\rangle$. Here, $\mathcal{S}$ denotes the state space, where $\bm{s} \in \mathcal{S}$ corresponds to the state representation derived from UAV observations. $\mathcal{A}$ represents the action space, and $\bm{a} \in \mathcal{A}$ is a predefined action consisting of two components: action type $a_{\text{type}}$ and action range $a_{\text{range}}$, where $a_{\text{type}} \in \{ \text{eastward}, \text{rightward}, \text{northward}, \text{southward}, \text{down}, \text{up}, \text{stop} \}$ and $a_{\text{range}} \in \{ 1, 2, \dots, A \}$. They specify the movement direction and distance, respectively. The observation space is denoted by $\mathcal{O}$. The state transition function is represented as $T(\bm{s}'|\bm{s}, \bm{a})$. The observation function $\Omega(\bm{o}|\bm{s}, \bm{a})$ denotes the probability of receiving observation $\bm{o}$ after executing action $\bm{a}$ in state $\bm{s}$. $\gamma$ is the discount factor used in calculating rewards. $R(\bm{s}, \bm{a})$ is the reward function, which is defined as follows:

\begin{equation}
	\label{eq3}
	R(\bm{s}, \bm{a}) =
	\begin{cases}
		D_1 - \sigma \cdot a_{\text{range}}, & \text{if } flag = 1,\ reco = 1 \\
		-D_1 - \sigma \cdot a_{\text{range}}, & \text{if } flag = 1,\ reco = 0 \\
		-D_2 - \sigma \cdot a_{\text{range}}, & \text{if } flag = 0
	\end{cases}
\end{equation}

The reward takes different values under three distinct scenarios. The variable $flag = 1$ indicates the end of the current episode, whereas $flag = 0$ means the agent continues interacting with the environment. The variable $reco$ characterizes whether the target is correctly identified after the agent executes action $\bm{a}$ in state $\bm{s}$: $reco = 1$ if the target recognition succeeds, and $reco = 0$ if it fails. Note that the classifier only classifies the target slice cropped according to the target bounding box. The positive constants $D_1$ and $D_2$ represent the base reward and base motion cost, respectively. $\sigma$ denotes the movement cost coefficient: when the action type $a_{\text{type}} = \text{stop}$, $\sigma = 0$; otherwise, $\sigma$ is set to a fixed value $C$. A larger $\sigma$ indicates that the reward function places greater emphasis on the agent's motion cost. Based on the above settings, we can rewrite the policy optimization problem in Eq.~\ref{eq1} as follows:

\begin{equation}
	\label{eq4}
	\pi^* = \arg\max_{\pi \in \Pi} \mathbb{E}_{\bm{s} \sim d(\bm{s})} \left[ r(\bm{s}) \right]
\end{equation}

\begin{figure*}[!t]
	\centering
	\includegraphics[width=\linewidth]{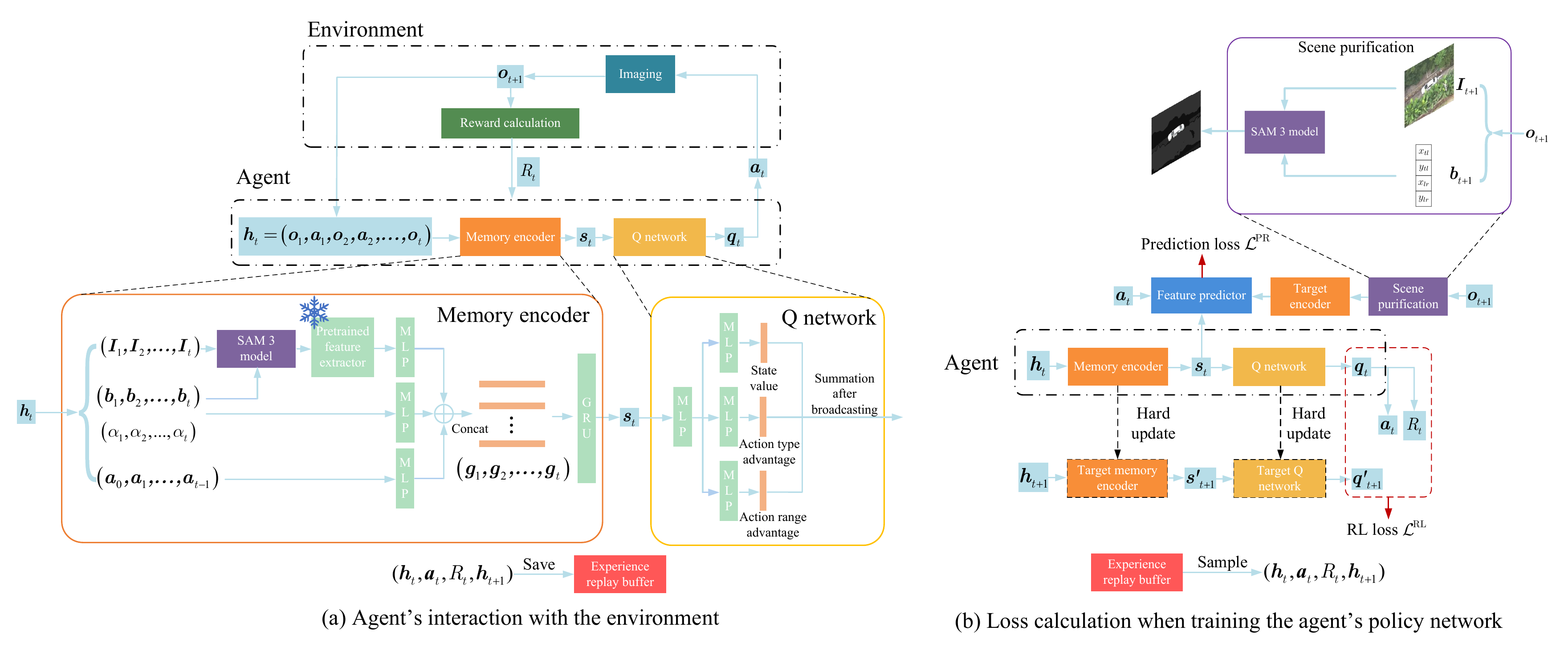}
	\caption{Flow chart of agent policy model-environment interaction and training loss calculation in the WMPL method.}
	\label{fig_network}
\end{figure*}

Here, $\Pi$ denotes the set of all feasible policies. The initial state distribution is given by $d(\bm{s})$. The return in an episode is represented by $r(\bm{s})$, which is defined as the weighted sum of rewards over $T_{\text{max}}$ time steps, with the mathematical expression as follows:

\begin{equation}
	\label{eq5}
	r(\bm{s}) := \mathbb{E}_{\substack{\bm{a}_t \sim \pi(\bm{a}_t|\bm{s}_t), \\ \bm{s}_{t+1} \sim T(\bm{s}_{t+1}|\bm{s}_t,\bm{a}_t)}} \left[ \sum_{t=0}^{T_{\text{max}}-1} \gamma^t R(\bm{s}_t, \bm{a}_t) \bigg| \bm{s}_0 = \bm{s} \right]
\end{equation}

It should be noted that in the POMDP, the aforementioned state $\bm{s}$ cannot be directly observed and must instead be estimated based on the observation history $\bm{h}$.

Furthermore, given that the agent's policy is trained and evaluated across different environments, we introduce the Contextual Markov Decision Process (CMDP) \cite{40}. CMDP is a variant of POMDP that incorporates contextual information, designed to assess the agent's zero-shot generalization (ZSG) capability when confronting new environments. Specifically, the CMDP is defined by the tuple $\left\langle \mathcal{S}', \mathcal{A}, \mathcal{C}, \mathcal{O}, T, \Omega, R, p, \gamma \right\rangle$, where $\mathcal{C}$ denotes the context space, $\mathcal{S}'$ represents the latent state space, and there exists a mapping $\mathcal{S}' \times \mathcal{C} \to \mathcal{S}$ that associates the latent state and context with the actual state space $\mathcal{S}$. The latent state variable $\bm{s}'$ explicitly specifies the relative positional relationship between the UAV and the target, while the context variable $\bm{c}$ further describes information such as the target's identity, appearance, and the spatial distribution of surrounding ground objects. Throughout an episode, the context variable remains constant.

For the CMDP with context space $\mathcal{C}$, the expected return of policy $\pi$ is calculated as: 

\begin{equation}
	\label{eq6}
	r(\pi, \mathcal{M} | \bm{c}) := \mathbb{E}_{\bm{c} \sim p(\bm{c})} \left[ r(\pi, \mathcal{M}_{\bm{c}}) \right].
\end{equation}

Here, $p(\bm{c})$ denotes the distribution of context variables in $\mathcal{C}$, and $\mathcal{M}_{\bm{c}}$ is the CMDP under the specific context variable $\bm{c}$. We define $\mathcal{C}_{\text{train}}$ and $\mathcal{C}_{\text{test}}$ as the training context space and test context space, respectively. By interacting with multiple training environments, the agent can learn the optimal policy $\pi^*_{\text{train}}$, which is defined as follows: 

\begin{equation}
	\label{eq7}
	\pi^*_{\text{train}} = \arg\max_{\pi \in \Pi} r(\pi, \mathcal{M} | \mathcal{C}_{\text{train}}).
\end{equation}

Finally, we evaluate the generalization capability of the policy in the test environments. Our ultimate goal is to solve for $\pi^*_{\text{train}}$ by improving the policy learning method, such that $r(\pi^*_{\text{train}}, \mathcal{M} | \mathcal{C}_{\text{test}})$ is maximized.

\subsection{World Model-aided Active Observation Policy Learning Method}
We leverage the world model constructed by AOD-JEPA to improve the quality of the agent’s state representations, thereby enhancing the policy learning performance of DRL algorithms. In practical applications, AOD-JEPA can be combined with any on-policy or off-policy reinforcement learning algorithm. In this paper, we perform joint training of the world model under the AOD-JEPA architecture and the Dueling DQN \cite{60} policy network employed in the MAP method \cite{30} , and we refer to this proposed AOD policy learning method as WMPL. Next, we elaborate on the detailed workflow of this method.

First, as illustrated in Fig.~\ref{fig_network}(a), the agent interacts with the environment multiple times and generates interaction data $({\bm{h}_t}, {\bm{a}_t}, R_t, {\bm{h}_{t+1}})$, which are stored in the experience replay buffer${\cal B}$. The observation history $\bm{h}_t$ consists of two components: the observation data sequence and the action sequence. To perform the AOD task, the agent is required to exploit a memory encoder $f_{\theta}$ to fuse the observation history and extract task-relevant state representations $\bm{s}_t$. Specifically, in the memory encoder adopted in this paper, we first leverage a multi-layer perceptron to process multi-modal information in the observation history chronologically, yielding diverse types of features corresponding to each time step. These features are then concatenated to form a feature sequence $({\bm{g}_1},{\bm{g}_2},…, {\bm{g}_t}) $, which is subsequently fused via a Gated Recurrent Unit network to generate the state representation $\bm{s}_t$. Notably, in the branch dedicated to processing observed images, we follow the practice in \cite{30} and adopt the backbone of the ResNet50 network \cite{62}, which is pre-trained on ImageNet \cite{61}, to process observed images and extract visual features. Next, the Q-network $Q_{\phi}$ outputs value estimates $\bm{q}_t$ for all candidate actions based on the abstracted state representations. Since each action comprises two components, namely action type and action magnitude, we first utilize two separate branches within the Q-network to evaluate the values of all possible action choices in these two dimensions respectively. Subsequently, through broadcasting and summation operations, we obtain the value estimates $\bm{q}_t$ for all action combinations on the plane spanned by the action type and action magnitude dimensions. During training, the action $\bm{a}_t$ is selected from $\bm{q}_t$ using an $\epsilon$-greedy strategy. The UAV platform then flies to a new observation position according to the selected action, and the environment feeds back a new observed image $\bm{o}_{t+1}$ and a reward $R_t$.

During the training phase illustrated in Fig.~\ref{fig_network}(b), when the amount of experience data stored in the experience replay buffer ${\cal B}$ exceeds the threshold $thre$, a batch of interaction experience data is randomly sampled from it and used for loss calculation. The loss to be computed during the policy network update consists of two components: the reinforcement learning loss and the feature prediction loss. In the Dueling DQN algorithm, to ensure the stability of policy network updates, target memory encoder ${\hat f}_{\theta}$ and target Q-network ${\hat Q}_{\phi}$ are usually introduced to estimate the target action values. During the training phase, the two target networks rewrite their own weights through a hard update mechanism, i.e., the weights of $f_{\theta}$ and $Q_{\phi}$ are assigned to ${\hat f}_{\theta}$ and ${\hat Q}_{\phi}$ every $N_u$ episodes. On this basis, the reinforcement learning loss can be formulated as:

\begin{equation}
	\label{eq8}
	\begin{array}{l}
		{{\cal L}^\text{RL}}(\theta ,\varphi ) = {\mathbb{E}_{({{\bm{h}}_t},{{\bm{a}}_t},{R_t},{{\bm{h}}_{t + 1}})\sim{\cal B}}}[Q(f({{\bm{h}}_t};\theta ),{{\bm{a}}_t};\varphi )\\
		\begin{array}{*{20}{c}}
			{}&{}&{}&{}
		\end{array} - ({R_t} + \gamma \mathop {\max }\limits_{{\bm{a}}'} \hat Q(\hat f({{\bm{h}}_{t + 1}};\hat \theta ),{\bm{a}}';\hat \varphi )){]^2}
	\end{array}
\end{equation}

The feature prediction loss originates from the next-step observation feature prediction task. As shown in the upper part of Fig.~\ref{fig_network}(b), AOD-JEPA incorporates a scene purification operation based on JEPA, integrating prior knowledge of the AOD task. Specifically, we utilize the SAM3 model to segment targets via bounding boxes and retain their original pixels. Meanwhile, since the SAM3 model can segment all object instances of the same category in an image according to language prompts, we assign the same grayscale value to the masks of ground objects of the same category. This preserves semantic information while filtering out their color and texture information, thereby further eliminating the interference of irrelevant information on state representation learning. Finally, the simplified image is processed by a pre-trained ResNet50 backbone to obtain improved target features for prediction. In the memory encoder, we also perform the scene purification operation on the observation image sequence in the observation history. The feature predictor is a three-layer fully connected network $l_{\psi}$, which predicts the features of the next-step observation image based on the input state representation ${\bm{s}_t}$ and action ${\bm{a}_t}$. Finally, the predicted values are compared with the improved target features, and the prediction loss ${{\cal L}^\text{PR}}$ is calculated, which is defined as:

\begin{equation}
	\label{eq9}
	{{\cal L}^\text{PR}}(\theta ,\psi ) \!=\! {\mathbb{E}_{({{\bm{h}}_t},{{\bm{a}}_t},{R_t},{{\bm{h}}_{t + 1}})\mspace{-2mu}\sim\mspace{-2mu}{\cal B}}}[\text{MSE}(l(f({{\bm{h}}_t};\theta ),{{\bm{a}}_t};\psi ),{f^p}({{\bm{o}}_{t + 1}}))]
\end{equation}

where MSE denotes the mean squared error. Accordingly, the overall optimization objective of the proposed method is formulated as:

\begin{equation}
	\label{eq10}
	\mathop {\min }\limits_{\theta ,\varphi ,\psi } {\mathbb{E}_{({{\bm{h}}_t},{{\bm{a}}_t},{R_t},{{\bm{h}}_{t + 1}})\sim{\cal B}}}[{{\cal L}^\text{RL}}(\theta ,\varphi ) + \zeta {{\cal L}^\text{PR}}(\theta ,\psi )]
\end{equation}

During the training phase, while the agent interacts with the environment, we sample a batch of experience data from the experience replay buffer ${\cal B}$ for loss calculation and backpropagation. The weights of the two loss terms in the optimization objective are adjusted via the coefficient $\zeta$. In the inference phase, only the memory encoder $f_{\theta}$ and the Q-network $Q_{\phi}$ are retained for action selection.

\section{Benchmark}
\label{Benchmark}

\subsection{Evaluation Task and Metrics}
Given that current research in the AOD field primarily
focuses on single-target scenarios, this paper first establishes
an evaluation benchmark for UGAOD policy learning methods
under single-target settings. The data utilized include local
target slice images and their corresponding annotations. Based
on the definition of the AOD problem, we define the specific
evaluation task as follows: An agent’s policy model is trained in given training environments using a specific AOD policy learning method, afterward, the trained policy runs AOD task inference in test environments; evaluation metrics quantify its test performance to compare different AOD policy learning methods

We adopt three seed-averaged quantitative metrics for evaluation: return value, recognition rate, and movement distance. The recognition rate represents the task success probability, defined as the ratio of successfully recognized episodes at the final observation step to the total episodes. The average movement distance quantifies the time and energy consumption of UAV aerial perception, which is calculated as the average flight distance across all episodes. As a comprehensive metric, the average return value balances recognition performance and movement cost. Considering that AOD tasks pursue optimal perception performance with minimal motion overhead, we take the average return value as the primary evaluation indicator, which is computed by averaging the cumulative returns of all test episodes.

\subsection{Implementation Details}
Several implementation details are clarified for consistent and fair evaluation. All UAV observations are constrained within the predefined regional range of the ATRNet-LUDO dataset, and episodes are terminated and discarded once the agent moves out of bounds, with no statistical inclusion. During training, environments and UAV initial positions are randomly sampled under controlled random seeds. For testing, test environments are cyclically selected, while initial observation positions are randomly initialized. Additionally, the initial position is resampled if the target is fully occluded and cannot be localized from the starting view. For a fair comparison, all baseline methods adopt identical training and testing episode settings and are fully trained to convergence.

On the other hand, the agent's perception module is identical across all baseline methods—i.e., they share a single target classifier—thereby ensuring consistency in their basic perceptual capabilities. For the sake of simplicity, we adopt the ResNet18 classification network as the classifier. A small subset of unoccluded local target patches is cropped from the observed images used for policy training to construct the training set dedicated to this classifier.

\section{Experiments}
\label{Experiments}
The purpose of this chapter is to comprehensively evaluate the performance of various AOD policy learning methods on the established benchmark, while verifying the effectiveness and advancement of the world model based on the AOD-JEPA architecture in enhancing policy generalization performance.

\subsection{Baselines}
The AOD policy learning methods involved in the evaluation include MTL \cite{63}, MAP \cite{30}, SSL-MAP \cite{13}, IBE-MAP \cite{39}, and WMPL. To ensure fair comparison, all these methods adopt the dual-branch deep network model illustrated in Fig.~\ref{fig_network}(a), where the two network branches are dedicated to outputting action type selection and action magnitude selection, respectively. In addition, we further compare two heuristic active observation policies: the random observation policy and the forward observation policy. Under the random observation policy, the agent randomly selects an action from the optional actions for output at each time step. In contrast, adopting the forward observation policy means that the agent selects the north direction on the current horizontal plane with a magnitude of 1 unit at each decision-making step.

On the other hand, to verify the effectiveness and advancement of AOD-JEPA in constructing a world model for UGAOD tasks, we compare it with world model training methods adopting other SSL architectures. Most representative baseline methods among them are designed based on SSL pretext tasks derived from paradigms such as contrastive learning \cite{55,56}, reward model \cite{65}, inverse model \cite{66}, forward model \cite{12}, and predictive coding \cite{57}. Contrastive learning methods learn effective state feature representations by constructing positive-negative sample pairs, corresponding to the JEA architecture. Specifically, for a given batch of observation history data, data augmentation techniques (e.g., random masking) are first applied to images; the augmented data are regarded as positive samples, while other historical data within the same batch are treated as negative samples. Subsequently, the InfoNCE loss function \cite{54} is optimized to maximize the similarity between positive sample features and anchor features in the representation space, while pushing negative sample features away from the anchor features. Methods based on reward model, forward model, and inverse model design distinct pretext tasks: they either predict the next-step reward or state based on the current state and the action executed by the agent, or infer the action taken by the agent between two consecutive states. The core idea of the predictive coding method is reconstructing the next-step observation data from the current state and action, which enables the agent to automatically build an internal representation of the environment. It is a practice of GA architecture. To ensure fair comparison, we adopt the MAP \cite{30} policy learning method as the baseline, and integrate it with world model training methods based on different SSL architectures. We then observe and compare their respective gains to the agent's active observation policy learning.

\subsection{Parameter configurations}
To ensure good reproducibility of the experimental results in this chapter, we provide the following specifications for the default configurations of key parameters of the model and learning algorithm. First, for the agent policy network model in Fig.~\ref{fig_network}, its input is a local target slice image with a resolution of $3\times300\times300$ pixels. The output action space of the policy network model consists of two dimensions: action type and action magnitude. The optional action types include seven categories, namely upward, downward, eastward, westward, southward, northward, and stop. The maximum action magnitude is set as $A=4$, with optional values of $1,2,\dots,A$. Each unit corresponds to 40 meters on the horizontal plane and 30 meters on the vertical axis.

During the training phase, the number of training epochs is set to 100. In each epoch, the agent interacts with the training environment for 3000 episodes and updates its policy network accordingly. After the completion of each training round, the agent is evaluated in the test environment for 3000 episodes to verify its performance. The discount factor $\gamma$ in the reward function is set to 0.9, while the variables $D_1$, $D_2$, and $C$ are configured as 0.5, 0.01, and 0.01, respectively. For non-heuristic policy learning methods, the agent takes random actions with a probability $\epsilon$ for environmental exploration. The initial value of this probability is set to 1, which decays exponentially to a minimum value of 0.01 as the number of training episodes increases. The capacity of the experience replay buffer $\mathcal{B}$ is set to 20000, with its threshold $thre$ configured as 5000. For the hard update operation, the update interval $N_u$ of the target networks is set to 100. The Adam optimizer is adopted to update the entire policy network during training, with the learning rate set to $1\times10^{-4}$ and the batch size $bs$ set to 512.

\subsection{Benchmarking results and analysis}
Prior to the formal evaluation of various UGAOD policy learning methods, we first tested the performance of multiple classical deep detection and recognition models on the ATRNet-LUDO dataset, aiming to enrich the existing evaluation benchmarks for UAV image detection and recognition tasks. On this basis, we focused on analyzing the impact of occlusion on existing detection and recognition methods to verify the importance of active perception applications. Specifically, the first part of the experiments covers two major tasks: panoramic multi-target detection and local patch recognition. For the detection task, we deployed Faster R-CNN \cite{67}, RetinaNet \cite{68}, SSD512 \cite{69}, YOLOX-L \cite{70}, and Deformable DETR \cite{71} under the MMDetection framework\cite{72}. 

We randomly sampled 10\% of the annotated panoramic images corresponding to collection regions 1--5 as the training set, with the remaining images serving as the easy test set; the sample set corresponding to regions 6--10 was designated as the hard test set. For the recognition task, we selected VGG16 \cite{73}, ResNet50 \cite{62}, ViT-B \cite{74}, Swin-B \cite{75}, and ConvNeXt \cite{76} as baseline methods based on the PyTorch Model Zoo library \cite{77}. In terms of dataset division, we also randomly sampled 10\% of the local patch images corresponding to collection regions 1--5 as the training set, with the remaining images used as the easy test set; the annotated patch set corresponding to regions 6--10 was defined as the hard test set.

\begin{figure}
	\centering
	\includegraphics[width=\linewidth]{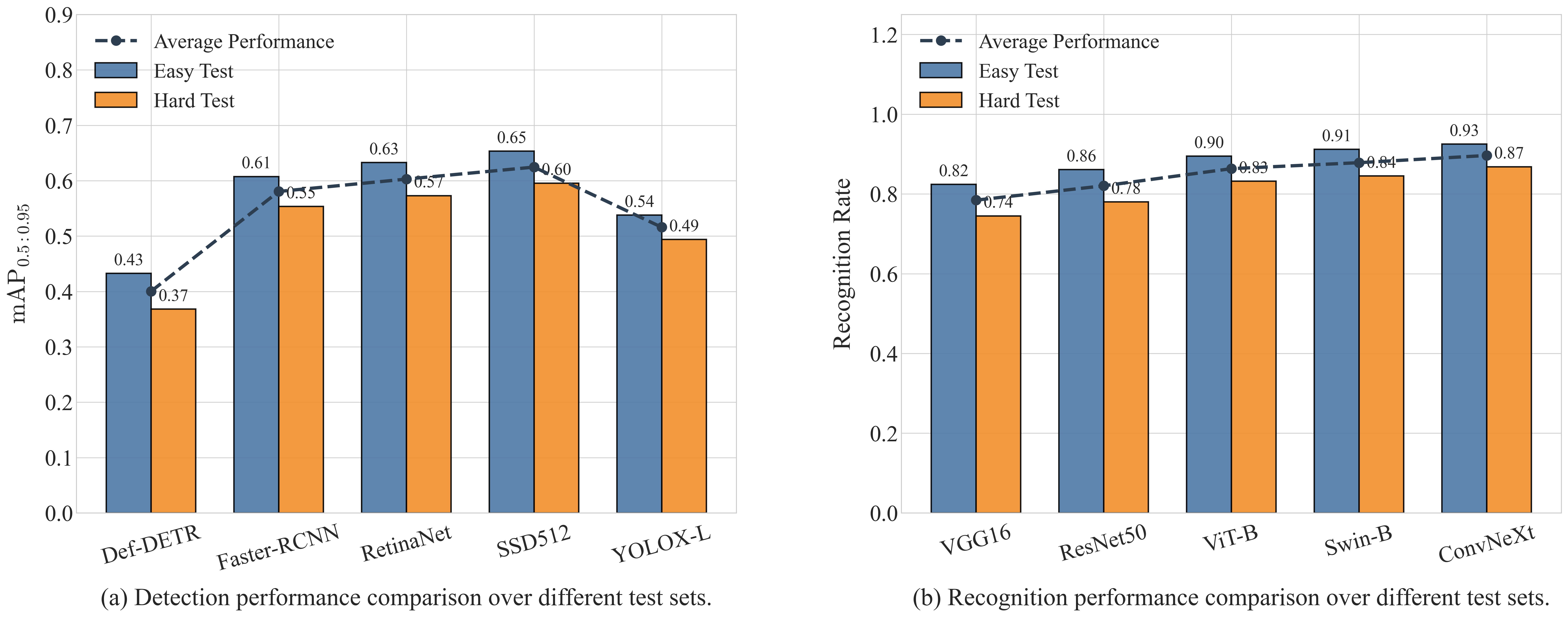}
	\caption{Evaluation results of generalization capability of classical detection and recognition methods on the ATRNet-LUDO dataset. (a) Performance comparison of five classical object detection methods on the easy and hard test sets;(b) Performance comparison of five classical object recognition methods on the easy and hard test sets.}
	\label{fig_14}
\end{figure}

Fig.~\ref{fig_14} quantitatively evaluates the generalization of mainstream detection and recognition models across easy and hard test splits. Detection performance is measured via $\text{mAP}_{0.5:0.95}$. Owing to multi-scale feature pyramid fusion, SSD512 and RetinaNet achieve competitive results of 0.65 and 0.63, respectively, though all detectors degrade by nearly five percentage points on the distribution-shifted hard test set. For recognition, classification accuracy varies substantially across architectures. Modern models such as ConvNeXt and Swin-B yield superior generalization, where ConvNeXt obtains accuracies of 0.93 and 0.87 on easy and hard sets, respectively. This verifies that large-kernel convolution designs can effectively extract robust fine-grained vehicle features for challenging UAV scenarios.

\begin{figure}
	\centering
	\includegraphics[width=\linewidth]{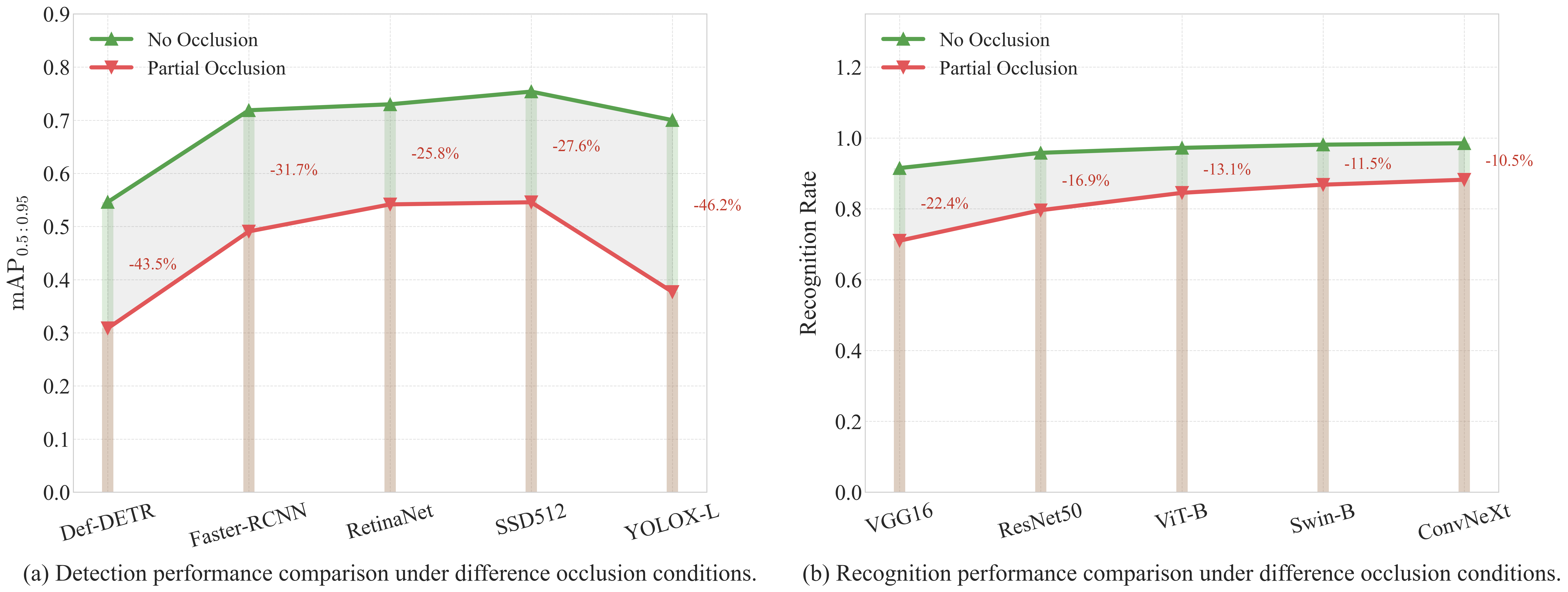}
	\caption{Evaluation results of classical detection and recognition methods on the easy test set under different occlusion conditions.}
	\label{fig_15}
\end{figure}

\begin{figure*}[!tb]
	\centering
	\includegraphics[width=\linewidth]{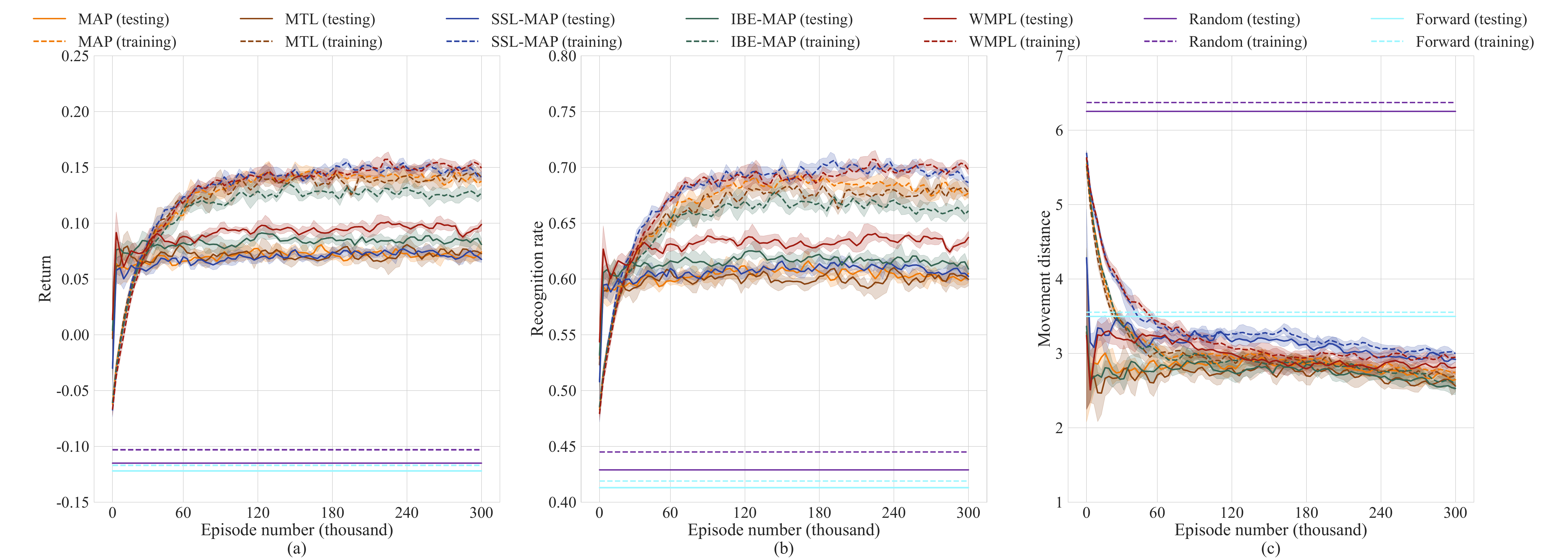}
	\caption{Comparison of training and testing performance curves between WMPL and other AOD policy learning methods under default settings. (a) Variations of training and testing returns with the increasing of training episodes; (b) Variations of training and testing recognition rates with the increasing of training episodes; (c) Variations of training and testing motion costs with the increasing of training episodes.}
	\label{fig_16}
\end{figure*}

\begin{table*}[!ht]
	\caption{Comparison of numerical results of different AOD policy learning methods under two dataset partition schemes}
	\label{tab2}
	\renewcommand\arraystretch{1.1}
	\centering
	\resizebox{0.7\linewidth}{!}{
		\begin{tabular}{cccccccc}
			\toprule
			\centering\multirow{3}{*}{Stage} & \centering\multirow{3}{*}{Method} & \multicolumn{3}{c}{Split 1} & \multicolumn{3}{c}{Split 2}\\ 
			\cmidrule(r){3-5}\cmidrule(r){6-8}
			~ & ~ & Return $\uparrow$  & Accuracy $\uparrow$ & \tabincell{c}{Movement \\ distance} $\downarrow$ & Return $\uparrow$  & Accuracy $\uparrow$ & \tabincell{c}{Movement \\ distance} $\downarrow$ \\
			
			\midrule
			\centering\multirow{7}{*}{Train} & MAP~\cite{30} & 0.146 & 0.685 & 2.708 & 0.184 & 0.733 & 2.700 \\
			~ & MTL~\cite{63} & 0.146 & 0.681 & 2.624 & 0.186 & 0.724 & \textbf{2.588} \\
			~ & SSL-MAP~\cite{13} & \textbf{0.155} & 0.702 & 2.954 & \textbf{0.192} & \textbf{0.747} & 2.832 \\
			~ & IBE-MAP~\cite{39} & 0.133 & 0.667 & \textbf{2.534} & 0.151 & 0.693 & 2.636 \\
			~ & WMPL & \textbf{0.155} & \textbf{0.703} & 2.894 & 0.185 & 0.741 & 2.854 \\
			~ & Random & -0.103 & 0.445 & 6.372 &  -0.087 & 0.467 & 6.380 \\
			~ & Forward & -0.117 & 0.419 & 3.552 & -0.09 & 0.460 & 3.590 \\
			\hline
			\centering\multirow{7}{*}{Test} & MAP~\cite{30} & 0.071&	0.600&	2.660 & 0.031 & 0.557 & 2.98 \\
			~ & MTL~\cite{63} & 0.069 & 0.594 & 2.566 & 0.038 & 0.558 & \textbf{2.598} \\
			~ & SSL-MAP~\cite{13} & 0.071 & 0.605 & 2.888 & 0.030 & 0.554 & 2.994 \\
			~ & IBE-MAP~\cite{39} & 0.086 & 0.615 & \textbf{2.488} & 0.044 & 0.570 & 3.068 \\
			~ & WMPL & \textbf{0.096} & \textbf{0.634} & 2.830 & \textbf{0.057} & \textbf{0.592} & 3.342 \\
			~ & Random & -0.115 & 0.429 & 6.252 & -0.138 & 0.400 & 6.270 \\
			~ & Forward & -0.122 & 0.413 & 3.498 & -0.161 & 0.355 & 3.450 \\
			
			\bottomrule
		\end{tabular}
	}
\end{table*}

We further analyze the occlusion-induced performance degradation of conventional detection and recognition models. Fig.~\ref{fig_15} compares the perception success rates under occlusion and non-occlusion conditions on the easy test set, where red and green triangles denote the performance under partial occlusion and clear observation, respectively, and their performance gap quantifies the occlusion interference. In Fig.~\ref{fig_15}(a), mainstream detectors including SSD512 and RetinaNet achieve high non-occlusion detection accuracy, with $\text{mAP}_{0.5:0.95}$ reaching 0.75 and 0.73, while occlusion causes severe performance deterioration. Specifically, YOLOX-L is most vulnerable to occlusion, suffering a 46.2\% accuracy drop from 0.70 to 0.38. For visual recognition, occlusion also produces prominent negative effects despite decent overall accuracy. VGG16 presents the poorest occlusion robustness with a 22.4\% performance decline. By contrast, ViT-B, Swin-B and ConvNeXt mitigate accuracy degradation via powerful global context modeling, yet they cannot fully eliminate occlusion interference. These results demonstrate that state-of-the-art passive perception methods still suffer universal performance limitations caused by occluded missing features. This manifests the necessity of active perception for robust visual recognition in complex scenarios.

To validate the generalization advantage of the proposed AOD-JEPA-based WMPL method, we compare it with existing AOD policy learning approaches under default settings, as summarized in Fig.~\ref{fig_16}. The plots illustrate the trends of return, recognition rate, and movement distance over training episodes, where solid and dashed lines correspond to performance on training and test environments, respectively. All experiments are repeated with five random seeds to eliminate DRL randomness, followed by sliding-window smoothing and mean averaging for stable visualization. The shaded regions denote result fluctuations, while only mean curves are plotted for random and forward policies due to their severe variance.

Experimental results show that both random and forward policies sustain low perception performance without observable improvement. By contrast, all learning-based active policies gradually enhance the UAV recognition accuracy with training, yet their performance is constrained by clear upper bounds. This limitation stems from abundant training environments that induce conflicting gradient updates and restrict optimization. Additionally, the evident training–testing performance gap indicates insufficient quality of the learned state representations. Overall, the proposed WMPL achieves the best generalization performance with the smallest generalization gap, yielding the highest recognition rate at low movement cost. Although MAP, MTL, and SSL-MAP attain competitive training returns, their performance degrades sharply during testing. While IBE-MAP maintains a moderate generalization gap, its low training sample efficiency leads to inferior final testing accuracy compared with WMPL.

Table~\ref{tab2} quantitatively benchmarks all policy learning methods under the two dataset partitioning protocols, with all metrics averaged over 30,000 inferences and bold values denoting the best performance. In the training phase, all DRL-based methods yield effective active observation policies that achieve high recognition accuracy with low motion cost. Nevertheless, all policies suffer severe recognition degradation in unseen test environments while maintaining stable movement overhead, leading to prominent generalization gaps in return performance. Notably, the second partitioning scheme produces a much larger generalization gap than the first. The first split only varies target layouts within similar background scenes, whereas the second split completely disjoints training and test scenarios, presenting higher generalization difficulty. Under both challenging settings, the proposed WMPL method consistently achieves superior test generalization performance, demonstrating its robustness for policy transfer across varying environmental distributions.

\begin{table}[!t]
	\caption{Performance comparison results of different policy learning methods with CLIP-pretrained visual encoders.}
	\label{tab3}
	\renewcommand\arraystretch{1.1}
	\centering
	\resizebox{0.8\linewidth}{!}{
		\begin{tabular}{ccccc}
			\toprule
			Stage & Method  & Return $\uparrow$  & Accuracy $\uparrow$ & \tabincell{c}{Movement \\ distance} $\downarrow$ \\
			\midrule
			\centering\multirow{7}{*}{Train} & MAP~\cite{30} & 0.152 & 0.695 & 2.862 \\
			~ & MTL~\cite{63} & 0.149 & 0.689 & \textbf{2.818} \\
			~ & SSL-MAP~\cite{13} & 0.155 & 0.702 & 2.954 \\
			~ & IBE-MAP~\cite{39} & 0.134 & 0.681 & 3.134 \\
			~ & WMPL & \textbf{0.159} & \textbf{0.712} & 3.084 \\
			~ & Random & -0.103 & 0.445 & 6.372 \\
			~ & Forward & -0.117 & 0.419 & 3.552 \\
			\hline
			\centering\multirow{7}{*}{Test} & MAP~\cite{30} & 0.078 & 0.611 & 2.842 \\
			~ & MTL~\cite{63} & 0.070 & 0.59 & \textbf{2.766} \\
			~ & SSL-MAP~\cite{13} & 0.071 & 0.605 & 2.888 \\
			~ & IBE-MAP~\cite{39} & 0.081 & 0.619 & 3.082 \\
			~ & WMPL & \textbf{0.095} & \textbf{0.636} & 3.008 \\
			~ & Random & -0.115 & 0.429 & 6.252 \\
			~ & Forward & -0.122 & 0.413 & 3.498 \\
			\bottomrule
		\end{tabular}
	}
\end{table}

\begin{table}[!t]
	\caption{Gain comparison of world model training with different SSL architectures on the MAP baseline policy learning method.}
	\label{tab4}
	\renewcommand\arraystretch{1.2}
	\centering
	\resizebox{\linewidth}{!}{
		\begin{tabular}{ccccc}
			\toprule
			Component & Method  & Return $\uparrow$  & Accuracy $\uparrow$ & \tabincell{c}{Movement \\ distance} $\downarrow$ \\
			\midrule
			\centering\multirow{7}{*}{w/o SP} & MAP~\cite{30} & 0.071 & 0.600 & \textbf{2.660} \\
			~ & Contrastive learning~\cite{55} & 0.065 (\textcolor{red}{-8.5\%}) & 0.597 & 2.832 \\
			~ & Reward model~\cite{65} & 0.072 (\textcolor{green}{+1.4\%}) & 0.603  & 2.814 \\	 	 
			~ & Inverse model~\cite{66} & -0.037 (\textcolor{red}{-152.1\%})  & 0.536 & 5.726 \\ 		 
			~ & Forward model~\cite{12} & 0.068 (\textcolor{red}{-4.2\%}) & 0.600 & 2.800 \\ 	 	 
			~ & Predictive coding~\cite{57} & 0.075 (\textcolor{green}{+5.6\%}) & 0.608 & 2.808 \\
			~ & AOD-JEPA & \textbf{0.083} (\textcolor{green}{+16.9\%}) & \textbf{0.623}  & 3.070 \\ 	 	 
			\hline
			\centering\multirow{7}{*}{w/ SP} & MAP~\cite{30} & 0.085 & 0.613 & \textbf{2.520} \\ 	 	 
			~ & Contrastive learning~\cite{55} & 0.077 (\textcolor{red}{-9.4\%}) & 0.608 & 2.742 \\  	 	 
			~ & Reward model~\cite{65} & 0.083 (\textcolor{red}{-2.4\%}) & 0.613 & 2.708 \\  	 	 
			~ & Inverse model~\cite{66} & -0.001 (\textcolor{red}{-101.2\%})  & 0.576 & 5.332 \\  		 
			~ & Forward model~\cite{12} & 0.086 (\textcolor{green}{+1.2\%}) & 0.617 & 2.652 \\  	 	 
			~ & Predictive coding~\cite{57} & 0.089 (\textcolor{green}{+4.7\%}) & 0.620 & 2.592 \\  	 	 
			~ & AOD-JEPA & \textbf{0.096} (\textcolor{green}{+12.9\%}) & \textbf{0.634} & 2.830 \\  	 	 
			\bottomrule
		\end{tabular}
	}
\end{table}

\begin{figure*}[!tb]
	\centering
	\includegraphics[width=0.9\linewidth]{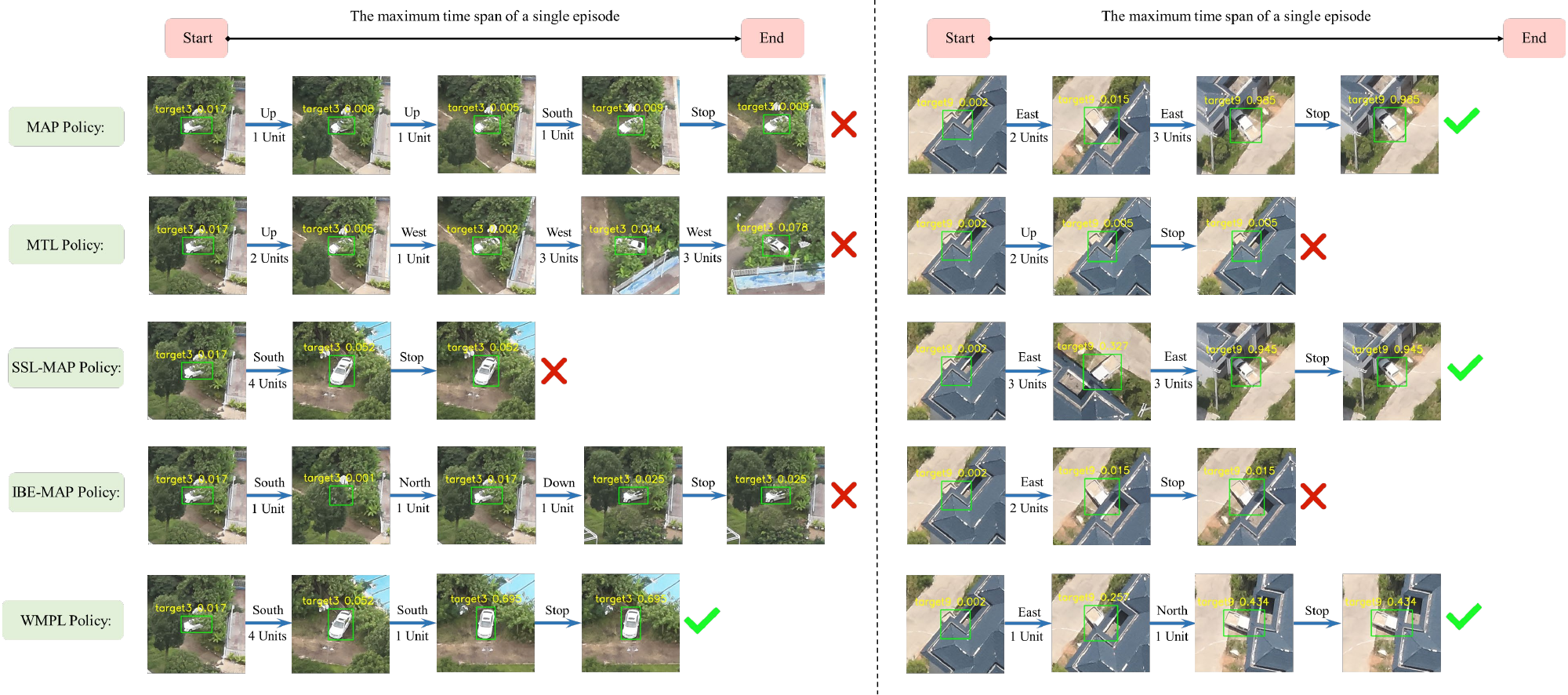}
	\caption{Demonstration of active observation sequence examples for UAVs adopting different policies in test environments.}
	\label{demo_seq}
\end{figure*}

On the other hand, the Contrastive Language-Image Pretraining (CLIP) visual encoder is equipped with powerful capability for extracting semantic information, which can help policy models achieve significant performance gains in embodied tasks such as object search and object rearrangement \cite{simplebut}. Therefore, we replaced the visual encoder in the policy network illustrated in Fig.~\ref{fig_network} with a ResNet50 backbone pretrained via the CLIP paradigm. On this basis, we trained the agent's active observation policies using various policy learning methods, and further compared the average performance results of different policies in both training and test environments. As shown in Table~\ref{tab3}, the numerical results indicate that with the aid of the AOD-JEPA architecture, WMPL still enables the agent to learn an active observation policy with the strongest generalization ability. Compared with the random or forward policy, it can significantly reduce the UAV's motion cost while bringing a target recognition rate gain of over 20\% in the test environment. In comparison with methods such as MAP, MTL, and SSL-MAP, WMPL can further improve the target recognition rate by an additional two percentage points or more in the test phase.

\begin{figure*}[!t]
	\centering
	\includegraphics[width=\linewidth]{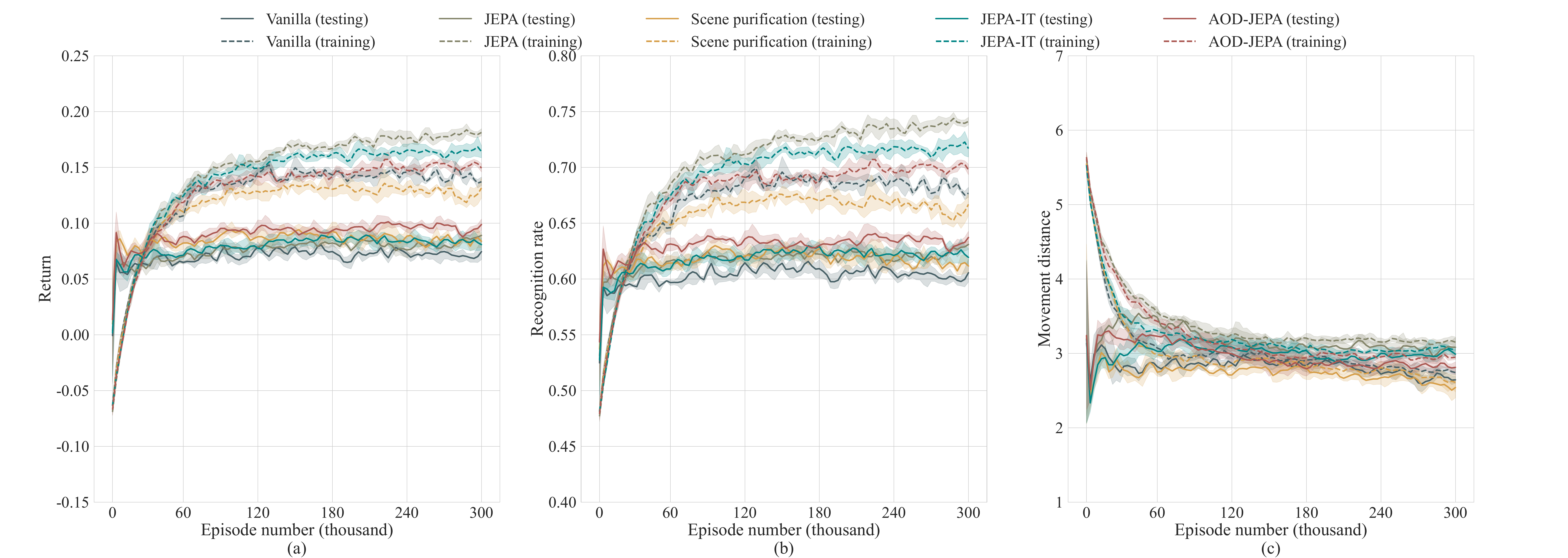}
	\caption{Comparison of training and testing performance curves of WMPL policy learning method under different ablation configurations.}
	\label{Abla_1}
\end{figure*}

Next, we compared the gains of world model training with different SSL architectures on the MAP baseline policy learning method. Table~\ref{tab4} reports the average results of policies learned by various methods over 30,000 inference runs in test environments. Contrastive learning, predictive coding, and AOD-JEPA correspond to the aforementioned SSL architectures JEA, GA, and JEPA, respectively. Unlike JEPA, the predictive pretext tasks of reward, inverse, and forward models adjust prediction model inputs or outputs based on the Markov decision process framework. In the table, ``w/o SP'' and ``w/ SP'' denote scenarios without and with scene purification, respectively. Overall, scene purification significantly boosts policy learning across all conditions and improves test-time policy returns. Among the SSL architectures, only predictive coding and AOD-JEPA stably enhance agent policy learning over the baseline, with AOD-JEPA delivering the most notable gains. We argue that predicting action-dependent next-observation changes from historical observations guides the model to focus on spatial relationships among the sensor platform, target, and ground objects. Furthermore, joint embedding prediction in the feature space and scene purification eliminate irrelevant information interference, improve state representation quality, and thus enhance policy generalization. In contrast, world model training with reward and forward models yields negligible benefits for policy learning, while contrastive learning-based and inverse model-based SSL architectures even impair performance. The core issue with contrastive learning lies in its instance discrimination paradigm: features of numerous batch negative samples (historical observations from different environments) are pushed away from anchor sample features. However, some negative samples share identical semantic information with anchors—i.e., the same action yields similar returns and subsequent states for both. This contradiction violates instance discrimination principles, undermining representation learning. The inverse model fails possibly because the two input variables of the prediction task have much higher dimensionality than the output, causing model collapse \cite{45, 78}. These results verify the effectiveness and superiority of AOD-JEPA for constructing the UGAOD task-specific world model.

To help readers more intuitively understand the differences between the active observation policies learned by various policy learning methods, we present the observation image sequences adopting different policies in the test environment in Fig.~\ref{demo_seq}. From the comparison of the demonstration examples of two episodes, the active observation policy via the WMPL method can achieve target recognition gains at a low motion cost. In contrast, under the same initial observation conditions, the policies corresponding to the other four policy learning methods face the problems of failing to find suitable viewpoints or incurring excessively high motion costs within the specified time. For instance, as shown in the first four rows on the left and the second and fourth rows on the right of the observation sequences, the observation images finally obtained by the agent through multiple decision-making steps cannot assist the perception module in accurately identifying the target. On the other hand, in the observation policies of the first and third rows of the right-hand episode, although the agent can capture observation images conducive to target recognition, it incurs a relatively high motion cost. In comparison, the policy corresponding to WMPL can efficiently improve the accuracy of target recognition at a lower motion cost, which reflects that the active observation policy learned based on the WMPL method has stronger generalization performance.

\subsection{Ablation Study}

To further verify the effectiveness of the AOD-JEPA architecture in the WMPL method, we conduct ablation experiments in this section. As shown in Fig.~\ref{Abla_1}, WMPL policy learning methods under different ablation configurations correspond to training and testing performance curves of different colors. Among them, Vanilla denotes the baseline MAP policy learning method without any improvements; JEPA refers to synchronously training a world model during policy network learning based on the JEPA architecture; and SP represents adding a scene purification operation to the baseline. IT in JEPA-IT is short for improved target, meaning scene purification is applied to the next-time observation image in the JEPA architecture while the prediction model input remains unchanged. AOD-JEPA integrates both joint embedding prediction and scene purification operations. From the results in Fig.~\ref{Abla_1}, both JEPA and scene purification significantly boost the agent's policy learning. Specifically, JEPA and JEPA-IT improve training sample efficiency, while scene purification eliminates irrelevant features to alleviate overfitting, thus restricting the rise of training performance curves to some extent. However, all these three methods enhance the generalization performance of active observation policies in the test environment and help the agent achieve higher test returns. A comparison of JEPA and JEPA-IT shows that purifying only the prediction target in JEPA yields negligible effects and fails to further improve policy generalization. In contrast, performing scene purification directly on the policy network input rapidly boosts generalization in early training, with a more notable effect than JEPA and JEPA-IT. Therefore, AOD-JEPA performs feature prediction on scene-purified observation images, which effectively improves the learned policy quality and greatly narrows the generalization gap between test and training performance curves.

\begin{table}[!t]
	\caption{Numerical evaluation results of agent-learned policies under different WMPL configurations in training and test environments.}
	\label{tab5}
	\renewcommand\arraystretch{1.1}
	\centering
	\resizebox{0.8\linewidth}{!}{
		\begin{tabular}{ccccc}
			\toprule
			Stage & Method  & Return $\uparrow$  & Accuracy $\uparrow$ & \tabincell{c}{Movement \\ distance} $\downarrow$ \\
			\midrule
			\centering\multirow{5}{*}{Train} & Vanilla & 0.146 & 0.685 & 2.708 \\
			~ & JEPA & \textbf{0.189} & \textbf{0.750} & 3.098 \\
			~ & SP & 0.132 & 0.666 & \textbf{2.558} \\
			~ & JEPA-IT & 0.172 & 0.727 & 3.062 \\
			~ & AOD-JEPA & 0.155 & 0.703 & 2.894 \\
			\hline
			\centering\multirow{5}{*}{Test} & Vanilla & 0.071 & 0.600 & 2.660 \\
			~ & JEPA & 0.083 (\textcolor{green}{+16.9\%}) & 0.623 & 3.070 \\
			~ & SP & 0.085 (\textcolor{green}{+19.7\%})& 0.613 & \textbf{2.520} \\
			~ & JEPA-IT & 0.082 (\textcolor{green}{+15.5\%}) & 0.622 & 3.070 \\
			~ & AOD-JEPA & \textbf{0.096} (\textcolor{green}{+35.2\%}) & \textbf{0.634} & 2.830 \\
			\bottomrule
		\end{tabular}
	}
\end{table}

\begin{figure}
	\centering
	\includegraphics[width=0.8\linewidth]{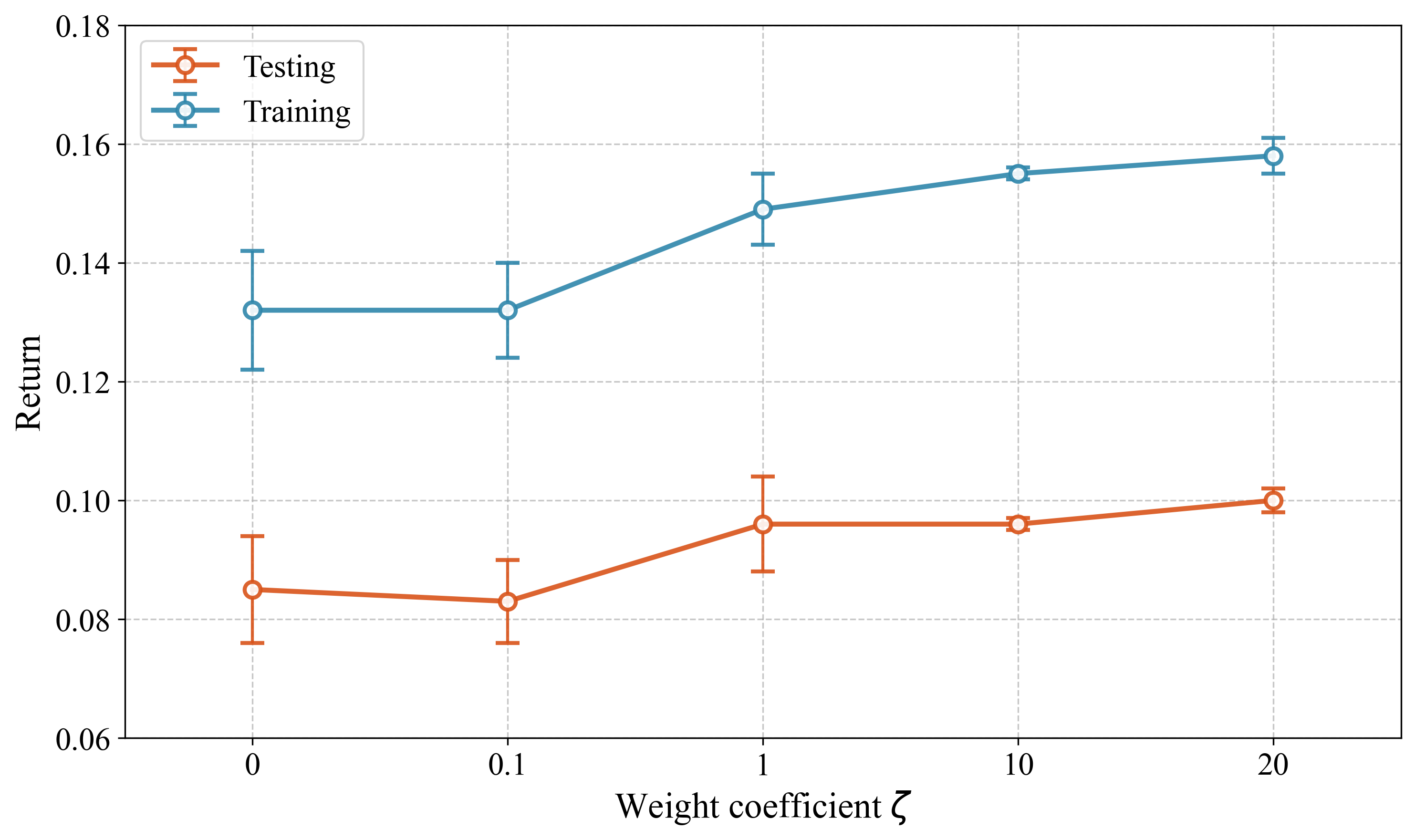}
	\caption{Line chart of training and testing evaluation results of WMPL policy learning method under different weight coefficients $\zeta$.}
	\label{Abla_2}
\end{figure}

The evaluation results of policy networks trained with AOD-JEPA under different configurations in training and test environments are presented in Table~\ref{tab5}. The comparison of these numerical results can clearly support the above conclusions. Regarding the main return metric, although the JEPA and scene purification operations have both positive and negative effects on policy learning in training environments, they all effectively improve the performance of active observation policies in test environments. while compared with the baseline version, JEPA and JEPA-IT can increase the target recognition rate by more than two percentage points in the test phase, and the scene purification operation can improve the recognition rate by 1.3 percentage points while reducing the motion cost. Ultimately, the fully-configured AOD-JEPA can increase the return level by 35.2\% from 0.071 of the baseline version, achieving a 3.4-percentage-point gain in target recognition rate at the cost of a slight increase in motion cost.

Finally, to further explore the impact of the proportion of the world model training loss in the total loss, we conduct an ablation experiment on the weight coefficient $\zeta$ in Eq.~(\ref{eq10}), and compare the training and testing performance of active observation policies under different values. The experimental results are shown in Fig.~\ref{Abla_2}. As $\zeta$ increases, the proportion of the world model training loss in the total loss gradually expands, bringing two evident gains to the agent's policy learning outcomes. First, the increase in $\zeta$ can improve the sample efficiency during training and enhance the generalization ability of the agent policy in the test environment; thus, both the training and testing returns of the corresponding policy are increased. On the other hand, since each data point in the figure is the average result of multiple runs with different random seeds, there is a certain degree of fluctuation, which we represent using line segments. It can be seen that with the increase in $\zeta$, not only do the mean values of the policy training and testing evaluation results rise, but their fluctuation ranges also gradually narrow, indicating that the performance becomes more stable. The above experimental phenomena demonstrate that the JEPA-based world model training can indeed help improve the state representation in the policy network and thus enhance its training and testing performance. Moreover, the larger the weight assigned during training, the heavier the constraint on state representation, and the more significant these beneficial improvements become.

\section{Conclusion and Future Perspectives}
\label{Conclusion}

In this paper, we release ATRNet-LUDO—the first large-scale real-world dataset for UGAOD tasks. It comprises 121,000 multi-view panoramic multi-target aerial images and 1.21 million multi-view local single-target aerial image patches, covering 10 vehicle categories across 40 scenarios. This dataset supports five UAV image-based perception tasks, including single/multi-target AOD, occluded target detection and recognition, and UAV visual target localization. Based on this dataset, we establish an evaluation benchmark for UAV-ground single-target AOD tasks, with explicit definitions of the evaluation task and metrics. This work facilitates the development and comparison of AOD policy learning methods, filling a critical gap in the field. On the methodological side, to tackle the poor generalization of active observation policies learned by existing methods, we propose the novel WMPL approach. Built on the AOD-JEPA architecture, WMPL constructs a UGAOD-specific world model to enhance state representation quality, thereby boosting the generalization performance of agent-learned policies.

UGAOD remains an emerging field with limited relevant research. As shown in our experiments, sample efficiency in training and policy generalization in testing are still compromised by conflicting update gradients across environments and suboptimal state representations, leaving ample room for improvement. We expect the proposed dataset and benchmark to lay a solid foundation for future research and help the academic community address these core challenges. In future work, we will explore the design and training of world model for UGAOD, enabling agents to learn more generalizable active observation policies through 3D scene understanding.

\footnotesize
\bibliographystyle{ieeetr}
\bibliography{IEEEabrv,ref}

\begin{thebibliography}{10}

\bibitem{1}
P.~K.~R. Maddikunta, S.~Hakak, M.~Alazab, S.~Bhattacharya, T.~R. Gadekallu,
  W.~Z. Khan, and Q.-V. Pham, ``Unmanned aerial vehicles in smart agriculture:
  Applications, requirements, and challenges,'' {\em IEEE sensors journal},
  vol.~21, no.~16, pp.~17608--17619, 2021.

\bibitem{2}
S.~Tao, Y.~Shengqi, L.~Haiying, G.~Jason, D.~Lixia, and L.~Lida, ``Mis-yolov8:
  An improved algorithm for detecting small objects in uav aerial photography
  based on yolov8,'' {\em IEEE Transactions on Instrumentation and
  Measurement}, 2025.

\bibitem{3}
J.~Dong, K.~Ota, and M.~Dong, ``Uav-based real-time survivor detection system
  in post-disaster search and rescue operations,'' {\em IEEE Journal on
  Miniaturization for Air and Space Systems}, vol.~2, no.~4, pp.~209--219,
  2021.

\bibitem{4}
J.~Xu, X.~Fan, H.~Jian, C.~Xu, W.~Bei, Q.~Ge, and T.~Zhao, ``Yoloow: A spatial
  scale adaptive real-time object detection neural network for open water
  search and rescue from uav aerial imagery,'' {\em IEEE transactions on
  geoscience and remote sensing}, vol.~62, pp.~1--15, 2024.

\bibitem{5}
B.~Yang, H.~Shi, and X.~Xia, ``Federated imitation learning for uav swarm
  coordination in urban traffic monitoring,'' {\em IEEE Transactions on
  Industrial Informatics}, vol.~19, no.~4, pp.~6037--6046, 2022.

\bibitem{6}
T.~M. Tran, D.~C. Bui, T.~V. Nguyen, and K.~Nguyen, ``Transformer-based
  spatio-temporal unsupervised traffic anomaly detection in aerial videos,''
  {\em IEEE Transactions on Circuits and Systems for Video Technology},
  vol.~34, no.~9, pp.~8292--8309, 2024.

\bibitem{33}
P.~Zhu, L.~Wen, D.~Du, X.~Bian, H.~Fan, Q.~Hu, and H.~Ling, ``Detection and
  tracking meet drones challenge,'' {\em IEEE transactions on pattern analysis
  and machine intelligence}, vol.~44, no.~11, pp.~7380--7399, 2021.

\bibitem{7}
D.~Cazzato, C.~Cimarelli, J.~L. Sanchez-Lopez, H.~Voos, and M.~Leo, ``A survey
  of computer vision methods for 2d object detection from unmanned aerial
  vehicles,'' {\em Journal of Imaging}, vol.~6, no.~8, p.~78, 2020.

\bibitem{8}
H.~Wang, C.~Wang, Q.~Fu, D.~Zhang, R.~Kou, Y.~Yu, and J.~Song, ``Cross-modal
  oriented object detection of uav aerial images based on image feature,'' {\em
  IEEE Transactions on Geoscience and Remote Sensing}, vol.~62, pp.~1--21,
  2024.

\bibitem{9}
P.~Mittal, R.~Singh, and A.~Sharma, ``Deep learning-based object detection in
  low-altitude uav datasets: A survey,'' {\em Image and Vision computing},
  vol.~104, p.~104046, 2020.

\bibitem{10}
F.~Li, X.~Li, Q.~Liu, and Z.~Li, ``Occlusion handling and multi-scale
  pedestrian detection based on deep learning: A review,'' {\em IEEE Access},
  vol.~10, pp.~19937--19957, 2022.

\bibitem{11}
J.~Aloimonos, I.~Weiss, and A.~Bandyopadhyay, ``Active vision,'' {\em
  International journal of computer vision}, vol.~1, no.~4, pp.~333--356, 1988.

\bibitem{13}
F.~Fang, W.~Liang, Y.~Wu, Q.~Xu, and J.-H. Lim, ``Self-supervised reinforcement
  learning for active object detection,'' {\em IEEE Robotics and Automation
  Letters}, vol.~7, no.~4, pp.~10224--10231, 2022.

\bibitem{21}
Z.~Zou, X.~Hu, and P.~Zhong, ``Active object detection for uav remote sensing
  via behavior cloning and enhanced q-network with shallow features,'' {\em
  IEEE Transactions on Geoscience and Remote Sensing}, 2025.

\bibitem{28}
P.~Ammirato, P.~Poirson, E.~Park, J.~Ko{\v{s}}eck{\'a}, and A.~C. Berg, ``A
  dataset for developing and benchmarking active vision,'' in {\em 2017 IEEE
  international conference on robotics and automation (ICRA)}, pp.~1378--1385,
  IEEE, 2017.

\bibitem{29}
X.~Han, H.~Liu, F.~Sun, and D.~Yang, ``Active object detection using double dqn
  and prioritized experience replay,'' in {\em 2018 International Joint
  Conference on Neural Networks (IJCNN)}, pp.~1--7, IEEE, 2018.

\bibitem{30}
X.~Han, H.~Liu, F.~Sun, and X.~Zhang, ``Active object detection with multistep
  action prediction using deep q-network,'' {\em IEEE Transactions on
  Industrial Informatics}, vol.~15, no.~6, pp.~3723--3731, 2019.

\bibitem{31}
S.~Liu, G.~Tian, X.~Shao, and S.~Liu, ``Behavior cloning-based robot active
  object detection with automatically generated data and revision method,''
  {\em IEEE Transactions on Robotics}, vol.~39, no.~1, pp.~665--680, 2022.

\bibitem{32}
W.~Ding, N.~Majcherczyk, M.~Deshpande, X.~Qi, D.~Zhao, R.~Madhivanan, and
  A.~Sen, ``Learning to view: Decision transformers for active object
  detection,'' in {\em 2023 IEEE International Conference on Robotics and
  Automation (ICRA)}, pp.~7140--7146, IEEE, 2023.

\bibitem{39}
X.~Jiang, T.~Liu, L.~Liu, Z.~Liu, and Y.~Liu, ``Uevavd: A dataset for
  developing uav's eye view active object detection,'' {\em IEEE Robotics and
  Automation Letters}, vol.~10, no.~6, pp.~6272--6279, 2025.

\bibitem{34}
Y.~Sun, B.~Cao, P.~Zhu, and Q.~Hu, ``Drone-based rgb-infrared cross-modality
  vehicle detection via uncertainty-aware learning,'' {\em IEEE Transactions on
  Circuits and Systems for Video Technology}, vol.~32, no.~10, pp.~6700--6713,
  2022.

\bibitem{35}
D.~Du, Y.~Qi, H.~Yu, Y.~Yang, K.~Duan, G.~Li, W.~Zhang, Q.~Huang, and Q.~Tian,
  ``The unmanned aerial vehicle benchmark: Object detection and tracking,'' in
  {\em Proceedings of the European conference on computer vision (ECCV)},
  pp.~370--386, 2018.

\bibitem{36}
I.~Bozcan and E.~Kayacan, ``Au-air: A multi-modal unmanned aerial vehicle
  dataset for low altitude traffic surveillance,'' in {\em 2020 IEEE
  International Conference on Robotics and Automation (ICRA)}, pp.~8504--8510,
  IEEE, 2020.

\bibitem{40}
D.~Ghosh, J.~Rahme, A.~Kumar, A.~Zhang, R.~P. Adams, and S.~Levine, ``Why
  generalization in rl is difficult: Epistemic pomdps and implicit partial
  observability,'' {\em Advances in neural information processing systems},
  vol.~34, pp.~25502--25515, 2021.

\bibitem{41}
R.~Kirk, A.~Zhang, E.~Grefenstette, and T.~Rockt{\"a}schel, ``A survey of
  zero-shot generalisation in deep reinforcement learning,'' {\em Journal of
  Artificial Intelligence Research}, vol.~76, pp.~201--264, 2023.

\bibitem{42}
J.~Lyu, L.~Wan, X.~Li, and Z.~Lu, ``Understanding what affects the
  generalization gap in visual reinforcement learning: Theory and empirical
  evidence,'' {\em Journal of Artificial Intelligence Research}, vol.~81,
  pp.~1--42, 2024.

\bibitem{45}
Y.~LeCun, ``A path towards autonomous machine intelligence.'' OpenReview
  Archive, 6 2022.
\newblock Direct Upload, Version 0.9.2.

\bibitem{43}
R.~Held and A.~Hein, ``Movement-produced stimulation in the development of
  visually guided behavior,'' {\em Journal of comparative and physiological
  psychology}, vol.~56, no.~5, p.~872, 1963.

\bibitem{47}
M.~Assran, Q.~Duval, I.~Misra, P.~Bojanowski, P.~Vincent, M.~Rabbat, Y.~LeCun,
  and N.~Ballas, ``Self-supervised learning from images with a joint-embedding
  predictive architecture,'' in {\em Proceedings of the IEEE/CVF Conference on
  Computer Vision and Pattern Recognition}, pp.~15619--15629, 2023.

\bibitem{58}
{Meta AI Research}, ``Sam 3: Segment anything with concepts.''
  \url{https://ai.meta.com/research/publications/sam-3-segment-anything-with-concepts/},
  2025.

\bibitem{37}
T.~Hodan, P.~Haluza, {\v{S}}.~Obdr{\v{z}}{\'a}lek, J.~Matas, M.~Lourakis, and
  X.~Zabulis, ``T-less: An rgb-d dataset for 6d pose estimation of texture-less
  objects,'' in {\em 2017 IEEE Winter Conference on Applications of Computer
  Vision (WACV)}, pp.~880--888, IEEE, 2017.

\bibitem{38}
Q.~Zhao, L.~Zhang, L.~Wu, H.~Qiao, and Z.~Liu, ``A real 3d embodied dataset for
  robotic active visual learning,'' {\em IEEE Robotics and Automation Letters},
  vol.~7, no.~3, pp.~6646--6652, 2022.

\bibitem{27}
Y.~Zhu, R.~Mottaghi, E.~Kolve, J.~J. Lim, A.~Gupta, L.~Fei-Fei, and A.~Farhadi,
  ``Target-driven visual navigation in indoor scenes using deep reinforcement
  learning,'' in {\em 2017 IEEE international conference on robotics and
  automation (ICRA)}, pp.~3357--3364, IEEE, 2017.

\bibitem{12}
D.~Jayaraman and K.~Grauman, ``End-to-end policy learning for active visual
  categorization,'' {\em IEEE transactions on pattern analysis and machine
  intelligence}, vol.~41, no.~7, pp.~1601--1614, 2018.

\bibitem{44}
A.~Dosovitskiy, G.~Ros, F.~Codevilla, A.~Lopez, and V.~Koltun, ``Carla: An open
  urban driving simulator,'' in {\em Conference on robot learning}, pp.~1--16,
  PMLR, 2017.

\bibitem{48}
A.~Bardes, Q.~Garrido, J.~Ponce, X.~Chen, M.~Rabbat, Y.~LeCun, M.~Assran, and
  N.~Ballas, ``Revisiting feature prediction for learning visual
  representations from video,'' {\em arXiv preprint arXiv:2404.08471}, 2024.

\bibitem{49}
M.~Abdelfattah and A.~Alahi, ``S-jepa: A joint embedding predictive
  architecture for skeletal action recognition,'' in {\em European Conference
  on Computer Vision}, pp.~367--384, Springer, 2024.

\bibitem{50}
C.~Bou~Chaaya, A.~M. Girgis, and M.~Bennis, ``Learning latent wireless dynamics
  from channel state information,'' {\em IEEE Wireless Communications Letters},
  vol.~14, no.~2, pp.~489--493, 2025.

\bibitem{51}
W.~Li, W.~Yang, T.~Liu, Y.~Hou, Y.~Li, Z.~Liu, Y.~Liu, and L.~Liu, ``Predicting
  gradient is better: Exploring self-supervised learning for sar atr with a
  joint-embedding predictive architecture,'' {\em ISPRS Journal of
  Photogrammetry and Remote Sensing}, vol.~218, pp.~326--338, 2024.

\bibitem{52}
W.~Li, W.~Yang, Y.~Hou, L.~Liu, Y.~Liu, and X.~Li, ``Saratr-x: Toward building
  a foundation model for sar target recognition,'' {\em IEEE Transactions on
  Image Processing}, vol.~34, pp.~869--884, 2025.

\bibitem{53}
A.~Riou, S.~Lattner, G.~Hadjeres, and G.~Peeters, ``Investigating design
  choices in joint-embedding predictive architectures for general audio
  representation learning,'' in {\em 2024 IEEE International Conference on
  Acoustics, Speech, and Signal Processing Workshops (ICASSPW)}, pp.~680--684,
  IEEE, 2024.

\bibitem{X-AnyLabeling}
CVHub, ``Advanced auto labeling solution with added features.''
  \url{https://github.com/CVHub520/X-AnyLabeling}, 2023.

\bibitem{60}
Z.~Wang, T.~Schaul, M.~Hessel, H.~Hasselt, M.~Lanctot, and N.~Freitas,
  ``Dueling network architectures for deep reinforcement learning,'' in {\em
  International conference on machine learning}, pp.~1995--2003, PMLR, 2016.

\bibitem{62}
K.~He, X.~Zhang, S.~Ren, and J.~Sun, ``Deep residual learning for image
  recognition,'' in {\em 2016 IEEE Conference on Computer Vision and Pattern
  Recognition (CVPR)}, pp.~770--778, 2016.

\bibitem{61}
J.~Deng, W.~Dong, R.~Socher, L.-J. Li, K.~Li, and L.~Fei-Fei, ``Imagenet: A
  large-scale hierarchical image database,'' in {\em 2009 IEEE Conference on
  Computer Vision and Pattern Recognition}, pp.~248--255, 2009.

\bibitem{63}
A.~Tavakoli, F.~Pardo, and P.~Kormushev, ``Action branching architectures for
  deep reinforcement learning,'' in {\em Proceedings of the aaai conference on
  artificial intelligence}, vol.~32, 2018.

\bibitem{55}
M.~Laskin, A.~Srinivas, and P.~Abbeel, ``Curl: Contrastive unsupervised
  representations for reinforcement learning,'' in {\em International
  conference on machine learning}, pp.~5639--5650, PMLR, 2020.

\bibitem{56}
J.~Zhu, Y.~Xia, L.~Wu, J.~Deng, W.~Zhou, T.~Qin, T.-Y. Liu, and H.~Li, ``Masked
  contrastive representation learning for reinforcement learning,'' {\em IEEE
  Transactions on Pattern Analysis and Machine Intelligence}, vol.~45, no.~3,
  pp.~3421--3433, 2022.

\bibitem{65}
R.~Rafailov, T.~Yu, A.~Rajeswaran, and C.~Finn, ``Offline reinforcement
  learning from images with latent space models,'' in {\em Learning for
  dynamics and control}, pp.~1154--1168, PMLR, 2021.

\bibitem{66}
C.~Allen, N.~Parikh, O.~Gottesman, and G.~Konidaris, ``Learning markov state
  abstractions for deep reinforcement learning,'' {\em Advances in Neural
  Information Processing Systems}, vol.~34, pp.~8229--8241, 2021.

\bibitem{57}
J.~Gornet and M.~Thomson, ``Automated construction of cognitive maps with
  visual predictive coding,'' {\em Nature Machine Intelligence}, vol.~6, no.~7,
  pp.~820--833, 2024.

\bibitem{54}
A.~van~den Oord, Y.~Li, and O.~Vinyals, ``Representation learning with
  contrastive predictive coding,'' {\em CoRR}, vol.~abs/1807.03748, 2018.

\bibitem{67}
S.~Ren, K.~He, R.~Girshick, and J.~Sun, ``Faster r-cnn: Towards real-time
  object detection with region proposal networks,'' in {\em Advances in Neural
  Information Processing Systems (NeurIPS)}, vol.~28, pp.~91--99, 2015.

\bibitem{68}
T.-Y. Lin, P.~Goyal, R.~Girshick, K.~He, and P.~Doll{\'a}r, ``Focal loss for
  dense object detection,'' in {\em Proceedings of the IEEE/CVF Conference on
  Computer Vision and Pattern Recognition (CVPR)}, pp.~2980--2988, 2017.

\bibitem{69}
W.~Liu, D.~Anguelov, D.~Erhan, C.~Szegedy, S.~Reed, C.-Y. Fu, and A.~C. Berg,
  ``Ssd: Single shot multibox detector,'' in {\em European Conference on
  Computer Vision (ECCV)}, pp.~21--37, 2016.
\newblock SSD512 is the extended version with input size 512$\times$512.

\bibitem{70}
Z.~Ge, S.~Liu, F.~Wang, Z.~Li, and J.~Sun, ``Yolox: Exceeding yolo series in
  2021,'' in {\em Proceedings of the IEEE/CVF Conference on Computer Vision and
  Pattern Recognition (CVPR) Workshops (NeurIPS 2021 Workshop)},
  pp.~3076--3085, 2021.

\bibitem{71}
X.~Zhu, W.~Su, L.~Lu, B.~Li, X.~Wang, and J.~Dai, ``Deformable detr: Deformable
  transformers for end-to-end object detection,'' in {\em Proceedings of the
  IEEE/CVF International Conference on Computer Vision (ICCV)}, pp.~303--312,
  2021.

\bibitem{72}
K.~Chen, J.~Wang, J.~Pang, Y.~Cao, Y.~Xiong, X.~Li, S.~Sun, W.~Feng, Z.~Liu,
  J.~Xu, {\em et~al.}, ``Mmdetection: Open mmlab detection toolbox and
  benchmark,'' in {\em Proceedings of the IEEE/CVF Conference on Computer
  Vision and Pattern Recognition (CVPR) Workshops}, pp.~68--77, 2019.

\bibitem{73}
K.~Simonyan and A.~Zisserman, ``Very deep convolutional networks for
  large-scale image recognition,'' {\em International Conference on Learning
  Representations (ICLR)}, 2015.
\newblock VGG16 is the 16-layer variant of the VGG architecture.

\bibitem{74}
A.~Dosovitskiy, L.~Beyer, A.~Kolesnikov, D.~Weissenborn, X.~Zhai,
  T.~Unterthiner, M.~Dehghani, M.~Minderer, G.~Heigold, S.~Gelly, {\em et~al.},
  ``An image is worth 16x16 words: Transformers for image recognition at
  scale,'' {\em International Conference on Learning Representations (ICLR)},
  2021.
\newblock ViT-B is the Base variant of the Vision Transformer.

\bibitem{75}
Z.~Liu, Y.~Lin, Y.~Cao, H.~Hu, Y.~Wei, Z.~Zhang, S.~Lin, and B.~Guo, ``Swin
  transformer: Hierarchical vision transformer using shifted windows,'' in {\em
  Proceedings of the IEEE/CVF International Conference on Computer Vision
  (ICCV)}, pp.~9992--10002, 2021.
\newblock Swin-B is the Base variant of the Swin Transformer.

\bibitem{76}
Z.~Liu, H.~Mao, C.-Y. Wu, C.~Feichtenhofer, T.~Darrell, and S.~Xie, ``A convnet
  for the 2020s,'' in {\em Proceedings of the IEEE/CVF Conference on Computer
  Vision and Pattern Recognition (CVPR)}, pp.~11976--11986, 2022.

\bibitem{77}
A.~Paszke, S.~Gross, F.~Massa, A.~Lerer, J.~Bradbury, G.~Chanan, T.~Killeen,
  Z.~Lin, N.~Gimelshein, L.~Antiga, {\em et~al.}, ``Pytorch: An imperative
  style, high-performance deep learning library,'' in {\em Advances in Neural
  Information Processing Systems (NeurIPS)}, vol.~32, pp.~8024--8035, 2019.

\bibitem{simplebut}
A.~Khandelwal, L.~Weihs, R.~Mottaghi, and A.~Kembhavi, ``Simple but effective:
  Clip embeddings for embodied ai,'' in {\em 2022 IEEE/CVF Conference on
  Computer Vision and Pattern Recognition (CVPR)}, pp.~14809--14818, 2022.

\bibitem{78}
A.~Dawid and Y.~LeCun, ``Introduction to latent variable energy-based models: a
  path toward autonomous machine intelligence,'' {\em Journal of Statistical
  Mechanics: Theory and Experiment}, vol.~2024, no.~10, p.~104011, 2024.

\end{thebibliography}

\begin{IEEEbiography}[{\includegraphics[width=1in,height=1.25in,clip]{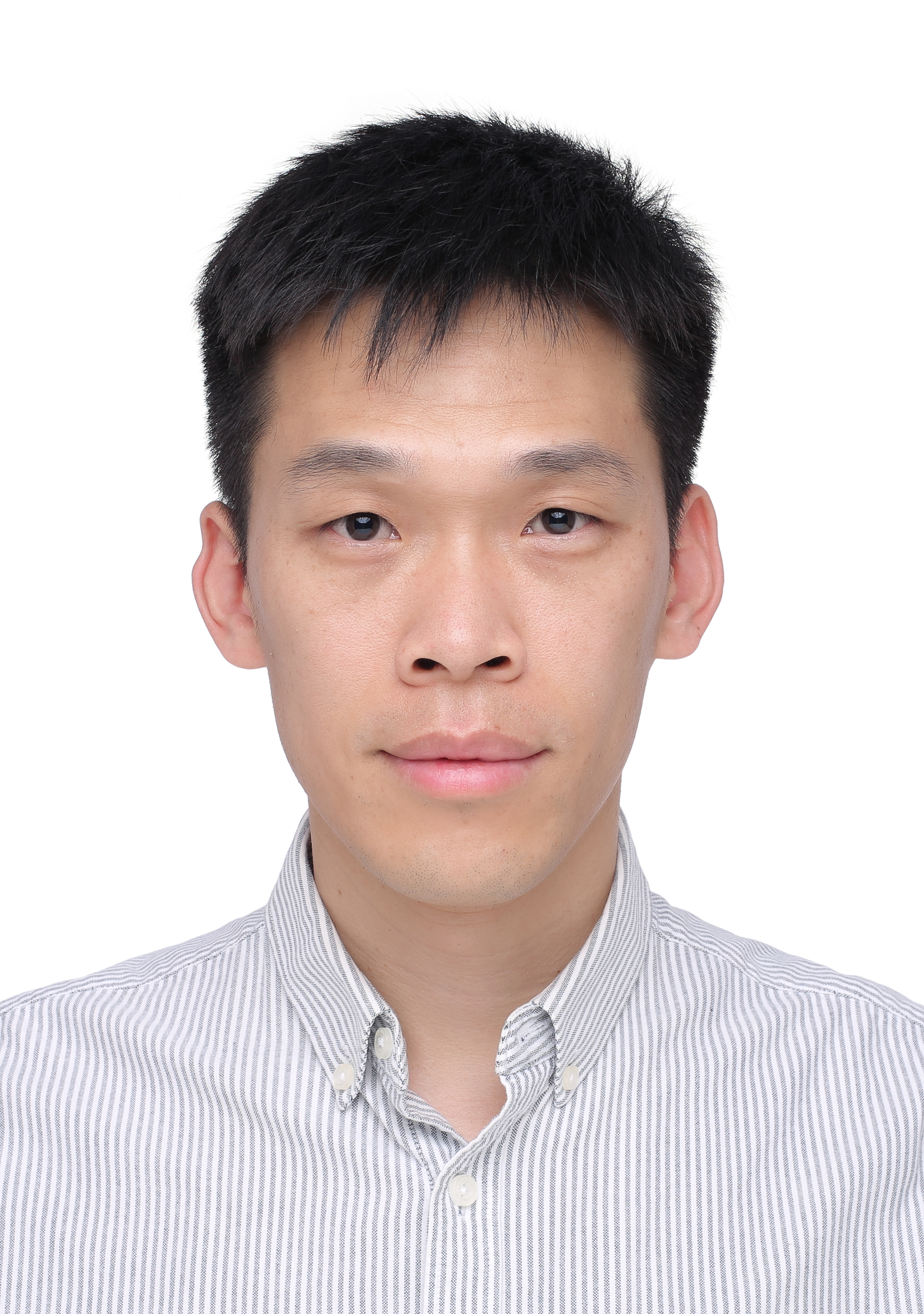}}]{Tianpeng Liu} received the B.E. and M.E and Ph.D. degrees from the National University of Defense Technology, Changsha, China, in 2008, 2011, and 2016 respectively. He is currently an associate professor at the College of Electronic Science and Technology. He has published numerous papers in respected journals, including IEEE Transactions on Aerospace and Electronic Systems and International Conference on Signal Processing. His primary research interests are radar signal processing, electronic countermeasure, and cross-eye jamming. 
\end{IEEEbiography}

\vspace{-2pt}
\begin{IEEEbiography}[{\includegraphics[width=1in,height=1.25in,clip]{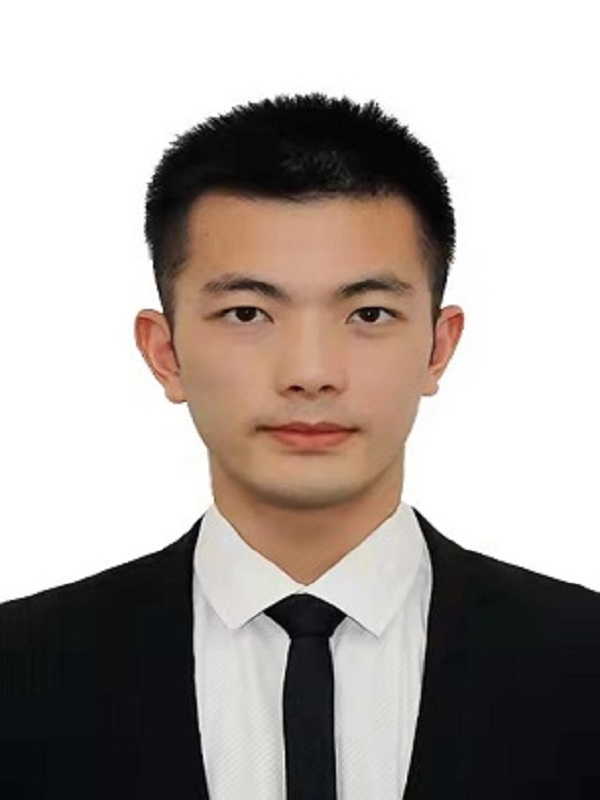}}]{Xinhua Jiang} received his Ph.D. degree from the National University of Defense Technology (NUDT), China, in 2026. He has published papers in respected journals, including IEEE Transactions on Circuits and Systems for Pattern Video Technology, IEEE Transactions on Geoscience and Remote Sensing, IEEE Journal of Selected Topics in Applied Earth Observations and Remote Sensing, IEEE Sensors Journal. His primary research interests are active vision, deep reinforcement learning, and representation learning.
\end{IEEEbiography}

\vspace{-2pt}
\begin{IEEEbiography}[{\includegraphics[width=1in,height=1.25in,clip]{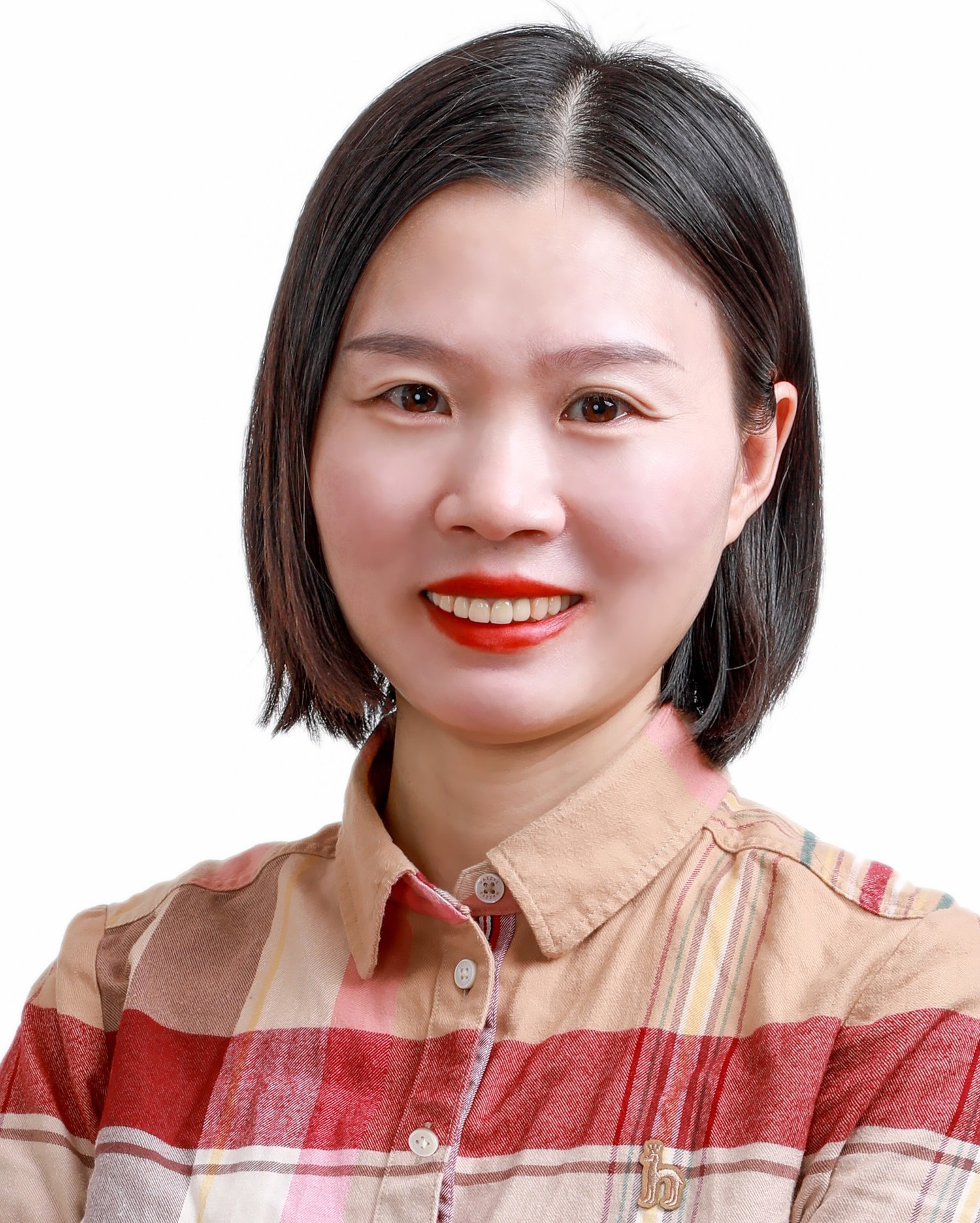}}]{Li Liu} received her Ph.D. degree in information and communication engineering from the National University of Defence Technology, China, in 2012. She joined the faculty at the National University of Defense Technology in 2012. Dr. Liu was co-chair of nine Internationa Workshops at several major venues, including CVPR, ICCV, and ECCV; she served as the leading guest editor of the special issues for IEEE TPAMI and IJCV. She currently serves as Associate Editor for IEEE Transactions on Circuits and Systems for Pattern Video Technology, IEEE Transactions on Geoscience and Remote Sensing, and Pattern Recognition. Her current research interests include computer vision, pattern recognition, and machine learning. Her papers currently have over 14000 citations in Google Scholar. She is a senior member of the IEEE.
\end{IEEEbiography}
\vspace{-2pt}
\begin{IEEEbiography}[{\includegraphics[width=1in,height=1.25in,clip]{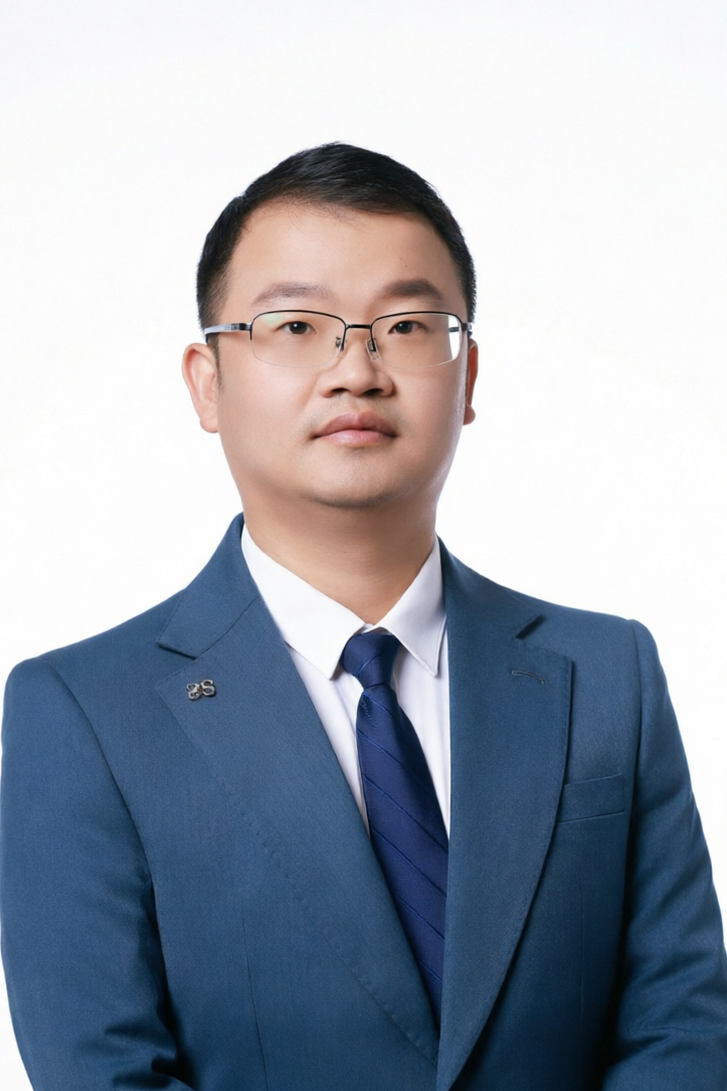}}]{Qinmu Shen} received the B.S. degree from Zhejiang University, Hangzhou, China, in 2005, and the M.S. and Ph.D. degree in mechanical engineering from National University of Defense Technology, Changsha, China, in 2008 and 2017 respectively. He was a Postdoctoral Researcher in information and communication engineering with National University of Defense Technology, Changsha, China. He is currently an asssociate professor in National University of Defense Technology. His research interests include intelligent target recognition, radar target recognition and multi-modal information fusion. 
\end{IEEEbiography}
\vspace{-2pt}
\begin{IEEEbiography}[{\includegraphics[width=1in,height=1.25in,clip]{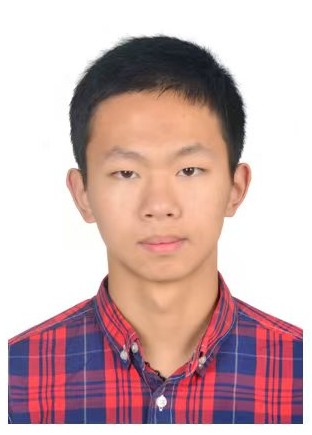}}]{Siwei Tang} received the B.S. degree in electronic information engineering from the National Univer sity of Defense Technology (NUDT), Changsha, China, in 2024, where he is currently pursuing the Ph.D. degree with the College of Electronic Science and Technology. His research interests include active perception and deep reinforcement learning.
\end{IEEEbiography}
\vspace{-2pt}
\begin{IEEEbiography}[{\includegraphics[width=1in,height=1.25in,clip]{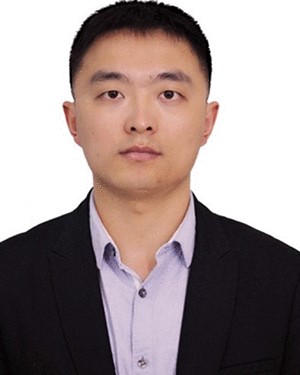}}]{Zhen Liu} received the Ph.D. degree in Information and Communication Engineering from National University of Defense Technology (NUDT), in 2013. He is currently a professor with the College of Electronic Science and Technology, NUDT. He has been awarded the Excellent Young Scientists Fund on his project titled “Intelligent Countermeasure for Radar Target Recognition” in 2020. His current research interests include radar signal processing, radar electronic countermeasure, compressed sensing, and machine learning.
\end{IEEEbiography}
\vspace{-2pt}
\begin{IEEEbiography}[{\includegraphics[width=1in,height=1.25in,clip]{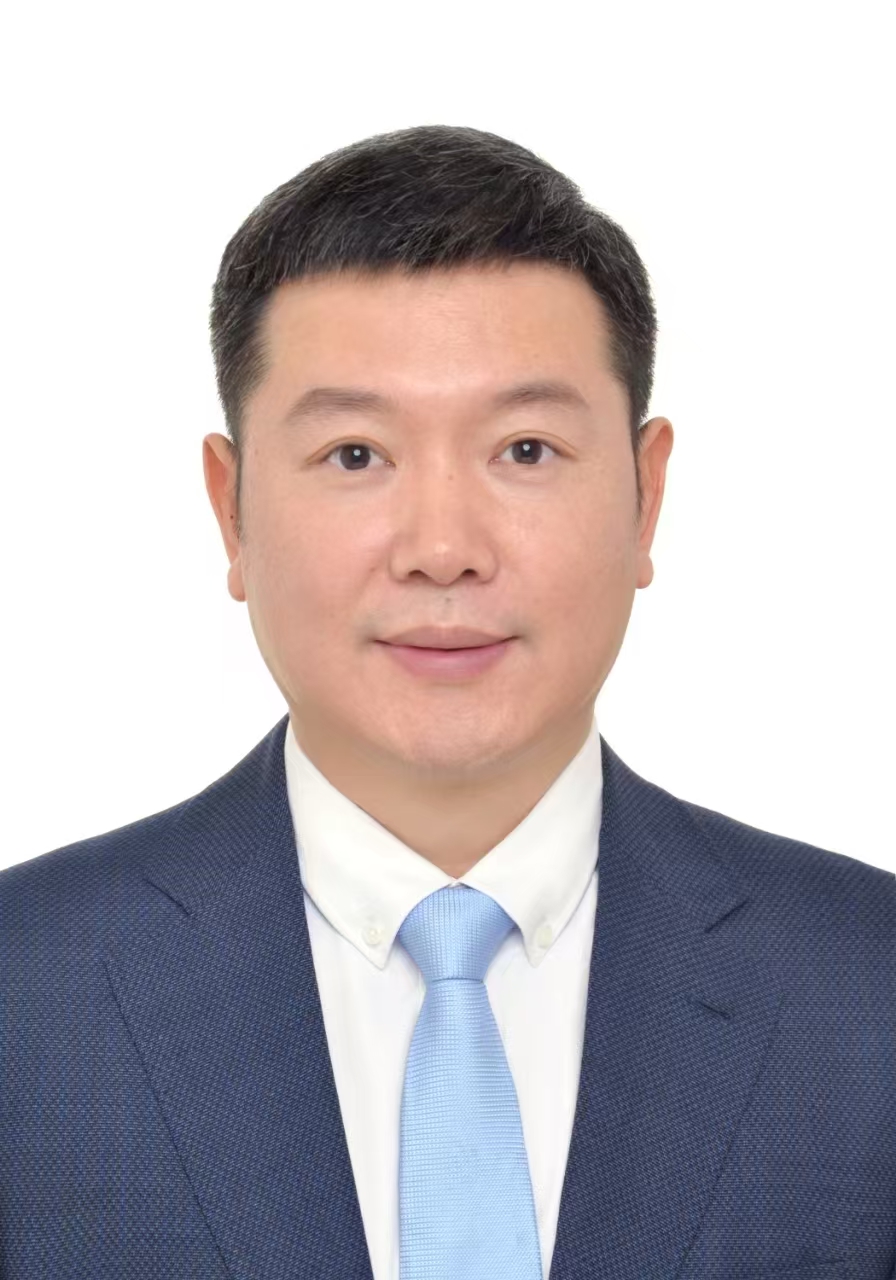}}]{Yongxiang Liu} received the B.S. and Ph.D. degrees from the College of Electronic Science at the National University of Defense Technology, Changsha, China, in 1997 and 2004, respectively. He is currently a Full Professor at the National University of Defense Technology. He has published numerous papers in respected journals, including IEEE Transactions on Image Processing, IEEE Transactions on Geoscience, and Remote Sensing. His research interests mainly include radar imaging, SAR image interpretation, and artificial intelligence. He is a member of the IEEE.
\end{IEEEbiography}

\end{document}